\documentclass[]{interact}

\usepackage{epstopdf}
\usepackage{subfigure}
\usepackage{parskip}
\usepackage{url}
\usepackage{float}
\usepackage{amssymb} 
\usepackage{amsmath}
\usepackage{graphicx}
\usepackage{hyperref}
\hypersetup{breaklinks=true}
\usepackage{epstopdf}
\usepackage{subfigure}
\usepackage{graphicx}
\usepackage{geometry} 
\usepackage{booktabs}
\usepackage{listings}
\usepackage{xcolor}
\usepackage{multirow}
\usepackage{enumitem} 
\theoremstyle{plain}

\usepackage{biblatex}

\theoremstyle{definition}

\theoremstyle{remark}

\begin{document}
\title{Evaluating Large Language Models on Spatial Tasks: A Multi-Task Benchmarking Study}

\author{
\name{Liuchang Xu\textsuperscript{a,b,d,+}, Shuo Zhao\textsuperscript{a,+}\thanks{\textsuperscript{+}These authors contributed equally to this work}, Qingming Lin\textsuperscript{a}, Luyao Chen\textsuperscript{a}, Qianqian Luo\textsuperscript{a}, Sensen Wu\textsuperscript{b}, Xinyue Ye\textsuperscript{c}, Hailin Feng\textsuperscript{a}, Zhenhong Du\textsuperscript{b,*} \thanks{CONTACT Zhenhong Du. Email: duzhenhong@zju.edu.cn} }
\affil{\textsuperscript{a}School of Mathematics and Computer Science, Zhejiang Agriculture and Forestry University, Hangzhou 311300, China; \textsuperscript{b}School of Earth Sciences, Zhejiang University, Hangzhou 310058, China; \textsuperscript{c}Department of Landscape Architecture and Urban Planning \& Center for Geospatial Sciences, Applications and Technology, Texas A\&M University, College Station, TX, 77843; \textsuperscript{d}Financial Big Data Research Institute, Sunyard Technology Co., Ltd., Hangzhou 310053, China}
}

\maketitle

\begin{abstract}
The advent of large language models such as ChatGPT, Gemini, and others has underscored the importance of evaluating their diverse capabilities, ranging from natural language understanding to code generation. However, their performance on spatial tasks has not been comprehensively assessed. This study addresses this gap by introducing a novel multi-task spatial evaluation dataset, designed to systematically explore and compare the performance of several advanced models on spatial tasks. The dataset encompasses twelve distinct task types, including spatial understanding and simple route planning, each with verified, accurate answers. We evaluated multiple models, including OpenAI’s gpt-3.5-turbo, gpt-4-turbo, gpt-4o, ZhipuAI’s glm-4, Anthropic’s claude-3-sonnet-20240229, and MoonShot’s moonshot-v1-8k, through a two-phase testing approach. Initially, we conducted zero-shot testing, followed by categorizing the dataset by difficulty and performing prompt tuning tests. Results indicate that gpt-4o achieved the highest overall accuracy in the first phase, with an average of 71.3\%. Although moonshot-v1-8k slightly underperformed overall, it surpassed gpt-4o in place name recognition tasks. The study also highlights the impact of prompt strategies on model performance in specific tasks. For example, the Chain-of-Thought (CoT) strategy increased gpt-4o’s accuracy in simple route planning from 12.4\% to 87.5\%, while a one-shot strategy enhanced moonshot-v1-8k’s accuracy in mapping tasks from 10.1\% to 76.3\%.
\end{abstract}

\section{Introduction}
The advent of large language models (LLMs) has revolutionized various domains\cite{zhou2024comprehensive}\cite{azaria2022chatgpt}\cite{kocon2023chatgpt}\cite{zhao2023survey}. Since the release of ChatGPT in late 2022, followed by other advanced models like gpt-4 and Google's Gemini\cite{team2023gemini}, these technologies have demonstrated significant potential in specialized technical fields. Of particular interest is their capability to handle complex spatial reasoning and geographic knowledge processing tasks. These models, trained on extensive datasets with billions of parameters, have shown promising results in natural language understanding, reasoning, and specialized domain applications, including geospatial analysis. Some advanced models, notably those with multimodal capabilities, can process both textual and visual geographic information, opening new possibilities for GIS applications\cite{roberts2023gpt4geo}\cite{li2023autonomous}\cite{app14167091}\cite{manvi2024geollm}\cite{juhasz2023chatgpt}.

Recent systematic performance comparisons of LLMs have revealed several key insights about their capabilities across various domains. In language processing, researchers have evaluated their performance on Chinese reading comprehension tasks\cite{kuo2024gpt}; in task planning, studies have examined LLMs' abilities in natural language planning and sequential decision-making\cite{zheng2024natural}. These evaluations, conducted using or referencing various comprehensive benchmarks, focus on three critical aspects: First, foundational knowledge and reasoning abilities have been extensively tested through benchmarks like C-Eval and AGIEval. The results indicate that while the best-performing models achieve high accuracy in basic knowledge tasks (\textgreater80\%), they face significant challenges in professional-level reasoning tasks, with accuracy rates dropping to 50-60\%\cite{huang2024c}\cite{zhong-etal-2024-agieval}. Second, practical application capabilities have been assessed through SuperCLUE and MMLU. These evaluations demonstrate that the best-performing models perform notably better in structured tasks compared to open-ended applications, with accuracy rates differing by 15-20\%\cite{xu2023superclue}\cite{li-etal-2024-cmmlu}. Third, comprehensive multi-dimensional evaluations through platforms like OpenCompass have shown that the best-performing models excel in language understanding and knowledge retrieval (\textgreater80\% accuracy). However, their performance declines significantly in tasks requiring advanced reasoning or domain-specific expertise, with accuracy rates typically falling below 65\%\cite{borji2023battle}. These systematic evaluations reveal a clear pattern: while LLMs demonstrate remarkable abilities in structured, knowledge-based tasks, even the best-performing models face notable challenges in complex reasoning and specialized domain applications\cite{lin2024wildbench}\cite{koubaa2023gpt}. 

\begin{figure}
\centering
\includegraphics[width=1\linewidth]{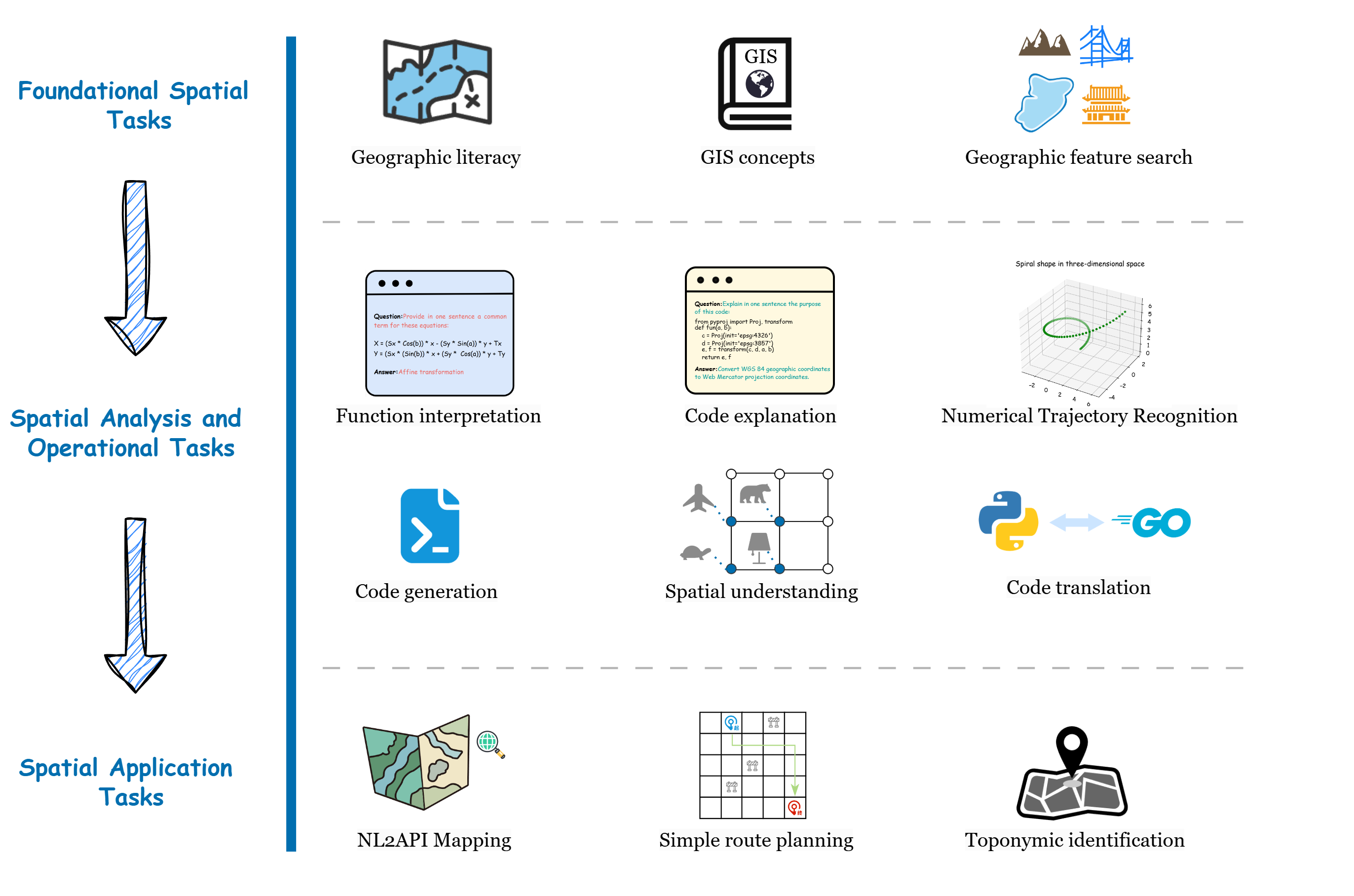}
\caption{\label{figure-1:PNG}Construction of task categories in spatial task datasets. }
\end{figure}

Alongside their general capabilities, there has been an increasing focus on the potential of LLMs to handle tasks within geographic information systems (GIS), particularly in spatial reasoning and geographic knowledge processing. Recent research has progressively revealed the significant potential of LLMs across various GIS tasks, highlighting both their promising capabilities and ongoing challenges. The foundational study by Peter Mooney et al. laid an important groundwork for evaluating ChatGPT's performance in GIS standardized exams. The study found that GPT-3.5 and GPT-4 achieved D and B+ grades, respectively, in an introductory GIS exam. This study not only provided a deep analysis of the strengths and limitations of LLMs in spatial concept understanding, but also offered valuable insights for educators and researchers on the application of AI models in specialized fields\cite{mooney2023towards}. However, as research has advanced, further studies have expanded upon these findings, illustrating both the rapid development and the persistent hurdles LLMs face when applied to complex spatial problems. The body of work in this area can be categorized into three key areas: spatial perception and understanding, spatial reasoning and analysis, and practical applications. Each of these areas reflects the broader progress and ongoing challenges that shape the current state of LLMs in GIS tasks. 
\begin{itemize}[leftmargin=2em, rightmargin=2em]
\item In the realm of spatial perception and understanding, significant progress has been made, particularly in cartography and spatial visualization. Tao and Xu’s pioneering research in map creation demonstrated that LLMs like ChatGPT could lower the barriers to map production and improve efficiency in cartographic tasks. However, their study also highlighted important challenges related to quality control, which indicates a need for further refinement in map generation processes\cite{tao2023mapping}. Subsequent advancements, such as those by Tao and Xu with GPT-4V, have expanded our understanding of LLMs' capabilities in map interpretation and spatial analysis\cite{xu2024map}. These studies have revealed that LLMs can effectively extract valuable information from a range of map types and perform complex quantitative spatial analyses. However, challenges remain, particularly in dealing with more intricate and diverse map types, where LLMs may struggle with map complexity and finer details. This highlights the need for future work focused on improving the models’ ability to handle these more complex scenarios. In a complementary study, Liu et al. examined LLMs’ capacity to process geographic diversity by analyzing DBpedia abstracts, further enhancing our understanding of how LLMs represent and interpret geographic knowledge\cite{liu2024measuring}. Their findings contribute to the ongoing development of LLMs as powerful tools for geographic knowledge representation, revealing both the strengths and the limitations of LLMs in capturing the diversity of geographic concepts. 
\item Research on spatial reasoning and analysis has increasingly focused on complex spatial problem-solving tasks, particularly those related to navigation. Aghzal et al. demonstrated that LLMs, particularly gpt-4, when optimized with few-shot prompting and fine-tuned BART and T5 models, can effectively handle intricate route planning tasks that require obstacle avoidance in dynamic environments\cite{aghzal2024large}. This study underscores the potential of LLMs to perform in real-world, complex spatial scenarios, where the ability to adapt to changing conditions is critical. Expanding on this, Yamada et al. developed sophisticated natural language navigation tasks that assess spatial structure representation\cite{yamada2023evaluating}. Their work revealed that models like gpt-3.5-turbo and gpt-4 exhibit strong capabilities in understanding and reasoning about spatial relationships, even in highly complex and cluttered environments, suggesting that these models could play a pivotal role in applications such as autonomous navigation and GIS. Furthermore, Sharma et al. explored the advanced analytical capabilities of LLMs by evaluating multiple models' performance on both 3D robotic trajectory data and 2D directional tasks\cite{sharma2023exploring}. Their findings provide valuable insights into LLMs’ adaptability across various forms of spatial reasoning, highlighting both their strengths in processing different data formats and the challenges they face when dealing with real-time, dynamic spatial data. These studies collectively demonstrate the growing potential of LLMs in spatial reasoning tasks but also reveal the need for further refinement to improve their performance in more complex, unpredictable environments.
\item In terms of practical applications in spatial tasks, LLMs have demonstrated substantial promise in a variety of real-world scenarios, showcasing their ability to provide efficient and actionable solutions in domains requiring spatial reasoning. In the realm of geographic information extraction, Hu et al. evaluated the performance of ChatGPT and gpt-4 in accurately identifying geographic locations from disaster-related social media content, emphasizing their potential for real-time data processing in crisis situations, where the need for timely and accurate geographic insights is paramount\cite{hu2023geo}. Their work highlights the capacity of LLMs to extract valuable geographic information from unstructured text, showcasing a key strength in processing large-scale, dynamic datasets. In geographic question answering, Feng et al. introduced GeoQAMap, an innovative system that integrates LLMs with structured geospatial data to provide interactive map visualizations for answering geographic queries\cite{feng_et_al:LIPIcs.GIScience.2023.28}. This work further underscores the utility of LLMs in offering actionable geographic insights through natural language interfaces, which can enhance decision-making in various applications such as urban planning, disaster response, and environmental monitoring. Additionally, in location-based services, Feng et al. explored innovative prompt engineering techniques to improve point-of-interest (POI) recommendation tasks, unlocking new possibilities for personalized and context-aware location-based services\cite{feng2024move}. Their research demonstrates the adaptability of LLMs to provide personalized recommendations based on user preferences and contextual information, which could revolutionize fields like tourism, navigation, and retail.
\end{itemize}
 
To systematically evaluate these diverse spatial applications, Hochmair et al. conducted a comprehensive evaluation of multiple chatbots across various spatial tasks, establishing crucial benchmarks for assessing LLMs' spatial reasoning capabilities\cite{hochmair2024correctness}. Their work provided an initial, broad overview of LLMs performance on spatial tasks, laying the groundwork for later benchmarks in this field. To further evaluate, new benchmarks for different spatial tasks have emerged successively. Tang and Kejriwal proposed GRASP, a grid-based benchmark containing 16,000 environments to evaluate commonsense spatial reasoning abilities of LLMs\cite{tang2024graspgridbasedbenchmarkevaluating}. Their findings revealed that even advanced models like GPT-4o face challenges in achieving consistent satisfactory solutions in spatial planning scenarios. Li and Bi developed STBench, which assesses LLMs' spatio-temporal understanding across four dimensions: knowledge comprehension, spatio-temporal reasoning, accurate computation, and downstream applications\cite{li2024stbenchassessingabilitylarge}. Their evaluation of 13 LLMs showed promising results in knowledge comprehension and reasoning tasks, while highlighting areas for improvement through various prompting strategies. In the urban domain, Feng and Li introduced CityBench, an interactive simulator-based evaluation platform that tests LLMs' capabilities in urban perception-understanding and decision-making tasks across 13 global cities\cite{feng2024citybenchevaluatingcapabilitieslarge}. These benchmarks evaluate LLMs performance from various aspects, particularly in the context of GIS-specific tasks, providing insights into their strengths and limitations in the domain of spatial reasoning.

Our research builds upon prior work, particularly in evaluating the ability of LLMs to handle spatial tasks. The work of Hochmair et al. provides valuable insights into LLMs' capabilities in basic spatial reasoning. However, their study focuses mainly on the analysis of individual spatial tasks and lacks a systematic, hierarchical evaluation framework, which limits a comprehensive examination of the interrelationships between various spatial tasks. Specifically, Hochmair’s study only evaluates LLMs performance in basic spatial tasks such as GIS concepts and simple code operations\cite{hochmair2024correctness}. While the study provides valuable testing of a single task, its approach does not encompass more complex spatial analysis and application tasks. To address this gap, we propose a more detailed and multidimensional classification method for spatial tasks. Furthermore, we integrate Geographic Information Science (GIS) with Bloom's Taxonomy of cognitive theory to develop a multilayered evaluation framework. This framework classifies spatial tasks into three levels: basic spatial tasks, spatial analysis and operations tasks, and spatial application tasks. In the design of these task dimensions, we specifically include the evaluation of complex spatial tasks. While previous studies have demonstrated that LLMs excel in basic geographic information processing, map creation, and fundamental spatial reasoning, they also highlight significant challenges in complex spatial analysis and specialized applications. Moreover, these studies primarily focus on specific types of spatial tasks and lack systematic, comprehensive benchmarks for evaluating LLMs performance across a broader range of spatial tasks. This one-sided approach limits a comprehensive understanding of LLMs capabilities and their limitations across different spatial tasks. Our research fills this gap by proposing a more scientifically rigorous and clearly structured evaluation framework. By categorizing spatial tasks into foundational, analytical, and applicative levels, we are better able to capture performance differences across various tasks and cognitive levels. This multilayered evaluation framework provides a clearer perspective for comprehensively understanding LLMs performance in spatial tasks. By combining cognitive complexity with the unique demands of GIS applications, our study offers a more robust and comprehensive benchmark for evaluating LLMs performance in spatial tasks. This framework not only reveals LLMs' capabilities in foundational tasks but also explores their potential in complex tasks and applications, providing important references for the future automation of spatial tasks and the development of spatial intelligence applications.

Our study makes several key contributions to address these challenges:

\begin{itemize}[leftmargin=2em, rightmargin=2em]
\item We developed a comprehensive and independent spatial task dataset (see Figure \ref{figure-1:PNG} ), encompassing twelve distinct task categories: Geographic literacy, GIS concepts, Natural Language to API Mapping(NL2API Mapping), function interpretation, code explanation, code generation, code translation, toponymic identification, spatial understanding, numerical trajectory recognition, geographic feature search, and simple route planning. 
\item Our study tested several well-known models, including three from OpenAI: gpt-3.5-turbo, gpt-4-turbo-2024-04-09, and gpt-4o, as well as ANTHROPIC's claude-3-sonnet-20240229, moonshot AI's moonshot-v1-8k, and zhipuai's glm-4. Additionally, we designed a comprehensive set of test scripts, utilizing API calls and precise control parameter settings to ensure the rigor and reproducibility of the experimental process. 
\item This study conducted two rounds of testing: the first round was a zero-shot test to evaluate the models' initial performance without any prompt tuning. Based on the results of the first round, we classified and assessed the difficulty levels of the spatial task dataset. For spatial tasks where the models performed poorly, we conducted prompt strategy tuning tests. The prompt strategies used included One-shot, Combined Techniques Prompt, Chain of Thought (CoT), and Zero-shot-CoT. 
\item We developed comprehensive evaluation strategies and introduced a new metric, Weighted Accuracy (WA), to more intuitively observe the models' capabilities. 
\end{itemize}

\section{Spatial Tasks Dataset}

To develop a comprehensive, reliable, and challenging spatial task dataset, we implemented a structured and efficient dataset construction strategy. The process consisted of three key stages: dataset design, data collection, and expert validation.

\begin{figure}
\centering
\includegraphics[width=0.8\linewidth]{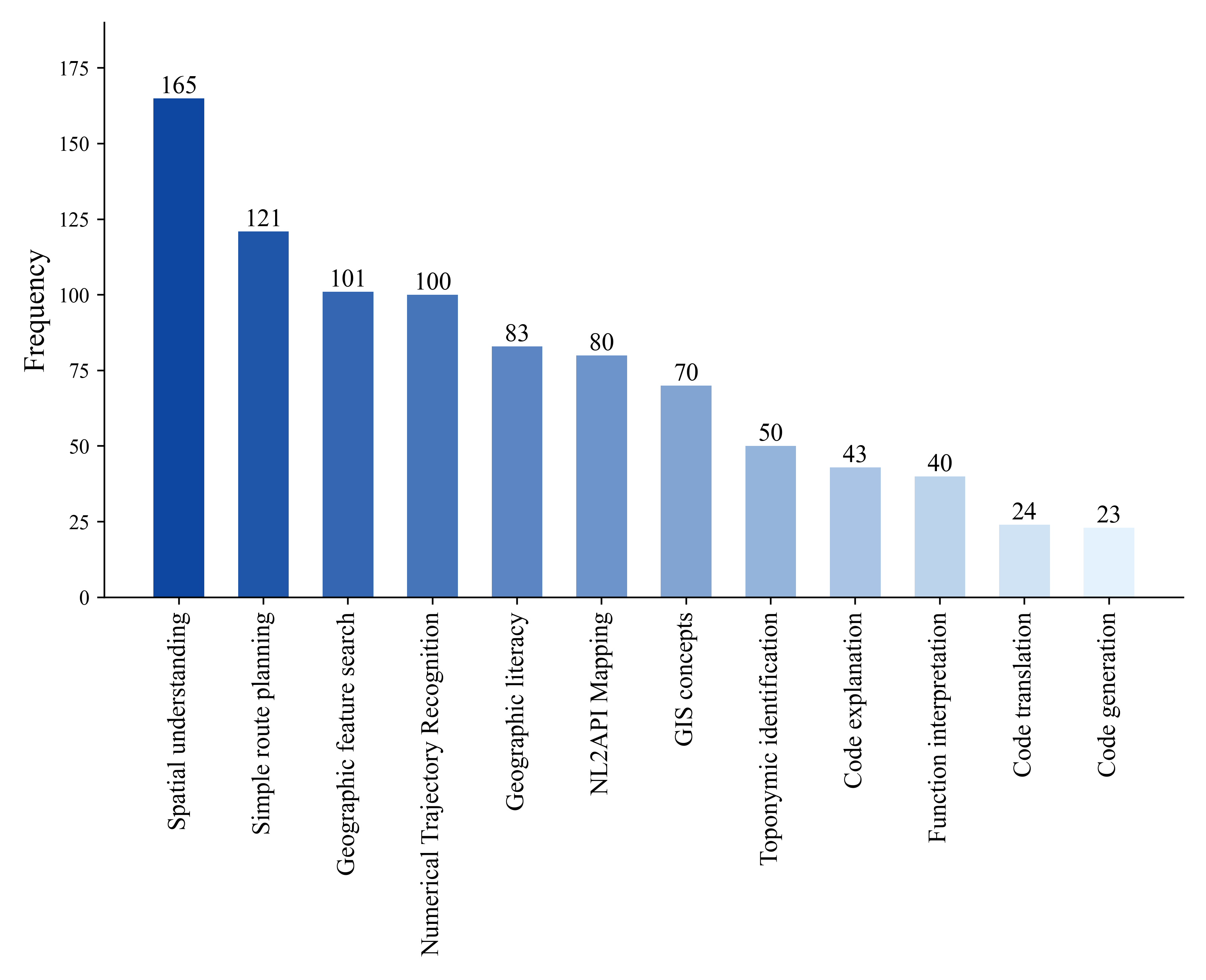}
\caption{\label{figure-2:PNG}Number of questions per category in the dataset (total number of questions: 900). }
\end{figure}

\subsection{Dataset Design}

Designing scientifically sound and reasonable dataset dimensions is crucial for building a high-quality spatial dataset. This study adopts an innovative approach by integrating expertise in geographic information science with cognitive theories from educational psychology to construct a multi-layered evaluation framework. Bloom’s (Benjamin Samuel Bloom) taxonomy of cognitive processes, one of the most influential theories in education, categorizes cognitive levels into knowledge, comprehension, application, analysis, synthesis, and evaluation. Drawing on this theoretical framework and incorporating the specific characteristics of the GIS-related tasks\cite{law2022getting}\cite{osullivan2010geographic}, this study divides the evaluation dimensions of the spatial dataset into three levels: foundational spatial tasks, spatial analysis and operational tasks, and spatial application tasks.

\subsubsection{Foundational Spatial Tasks}

Foundational spatial tasks form the cornerstone of spatial task evaluation, representing the essential cognitive and operational capabilities required for LLMs to process spatial information. These tasks are not only prerequisites for advanced spatial analysis but also critical for ensuring the quality of spatial data. Evaluating foundational spatial tasks allows for a comprehensive assessment of a LLMs's fundamental capabilities in spatial data collection, processing, and interpretation. Based on task characteristics and application requirements, we categorize foundational spatial tasks into three core subtasks: geographic feature search, geographic literacy, and GIS concepts.

\begin{itemize}[leftmargin=2em, rightmargin=2em]
\item Geographic Feature Search: This task primarily evaluates a system's ability to recognize and understand geographic entities and their attributes. It encompasses both natural geographic features (e.g., mountains, rivers, lakes) and human geographic elements (e.g., historical sites, cultural landmarks), including their identification and attribute extraction. In practical applications, this task requires the system to accurately extract geographic location information from textual descriptions and comprehend the political, economic, and cultural characteristics of these entities, as well as their spatial relationships. Studies indicate that current generative AI models exhibit certain limitations in representing and understanding geographic features, particularly at the local level, where significant regional disparities are observed. Moreover, their knowledge coverage of some geographic features remains inadequate\cite{liu2024measuring}. This suggests that, despite their potential, these models require further optimization to meet the demands of applications in GIS, tourism recommendations, and academic research.
\item Geographic Literacy: Geographic literacy tasks evaluate a system's comprehensive understanding of global geographic locations, topographic features, and demographic information. This includes basic knowledge of geographic elements such as cities, countries, major highways, river systems, and elevation data, as well as an understanding of topological relationships between geographic units. Geographic literacy is defined as the ability to understand, evaluate, and apply geographic knowledge, encompassing the identification of geographic facts (e.g., place names, spatial distributions, natural or man-made features) and the capacity to solve problems using spatial reasoning. This concept is inspired by spatial literacy studies, which focus on individuals' abilities to visualize, reason, and communicate spatial data, such as making inferences and decisions based on spatial relationships\cite{lane2019problematizing}. While spatial literacy and geographic literacy share some foundational abilities, geographic literacy places particular emphasis on integrating geographic knowledge with spatial operations. This distinction grants geographic literacy unique value in analytical tasks and practical applications. For instance, spatial literacy typically involves fundamental operations in two- and three-dimensional spaces, such as understanding adjacency relationships or constructing terrain models. In contrast, geographic literacy extends this by exploring how geographic knowledge (e.g., patterns of river system distribution or demographic characteristics) can be situated within broader geographic frameworks for analysis and reasoning. This integration of knowledge and reasoning enables geographic literacy to address complex geographic problems comprehensively, demonstrating significant utility in fields such as geographic information systems, environmental planning, and education.
\item GIS Concepts: GIS concept tasks focus on assessing a system’s understanding of the fundamental theories and operational methods of GIS. These tasks encompass core knowledge such as map projections, spatial data management, and analytical processing. They evaluate the system's ability to use GIS tools to solve complex geospatial problems through simulated scenarios and theoretical questions. As noted by Law and Collins in Getting to Know ArcGIS Desktop 10.8\cite{law2022getting}, GIS serves as a scientific framework for managing and analyzing data through geographic visualization and locational intelligence. O'Sullivan and Unwin, in Geographic Information Analysis\cite{osullivan2010geographic}, further emphasize that GIS analysis requires integrating spatial pattern recognition and statistical methods to scientifically model and predict complex geographic phenomena. Mastery of these foundational theories is crucial for ensuring the scientific rigor of subsequent spatial analyses.
\end{itemize}

\subsubsection{Spatial Analysis and Operational Tasks}

Spatial analysis and operational tasks form a pivotal component of geographic information systems and the evaluation framework, bridging foundational spatial cognition with practical application outcomes. These tasks assess not only a LLMs's capability to analyze spatial data but also its ability to transform analytical results into actionable operations. Based on functional characteristics and application requirements, we categorize spatial analysis and operational tasks into six subtasks: numerical trajectory recognition, spatial understanding, code explanation, function interpretation, code translation, and code generation.

\begin{itemize}[leftmargin=2em, rightmargin=2em]
\item Numerical Trajectory Recognition: Numerical trajectory recognition tasks focus on evaluating a system's ability to identify and analyze motion trajectories in two- or three-dimensional space. This includes recognizing directional changes (e.g., up, down, left, right) and describing shapes based on the arrangement of points (e.g., waveforms or straight lines). Such capabilities hold significant application value in areas like automated navigation systems, motion trajectory analysis, and GIS data processing, directly affecting system efficiency and accuracy. Research suggests that current LLMs demonstrate potential in handling numerical trajectory data and spatial reasoning tasks. For instance, Sharma evaluated the performance of models like ChatGPT-4 on 3D robotic trajectory data using the CALVIN dataset and achieved significant improvements through prefix prompting techniques\cite{sharma2023exploring}. This indicates that LLMs, by enhancing spatial reasoning capabilities, can provide robust support for geometric shape recognition and dynamic change analysis.
\item Spatial Understanding: Spatial understanding tasks focus on evaluating a system’s ability to represent and reason about complex spatial structures. These tasks involve natural language navigation challenges incorporating grids, hexagons, triangular grids, and circular or tree-like structures to test the system's comprehension and manipulation of spatial configurations. This spatial understanding is broadly applicable in fields such as autonomous driving and robotic route planning, requiring systems to not only identify physical locations but also perform abstract analysis and reasoning about spatial relationships. As noted by Yamada et al., LLMs such as gpt-4 and the Llama2 series demonstrate latent representational capabilities for spatial configurations like grids, circular arrangements, and tree structures when tackling natural language navigation tasks. However, their performance varies significantly across different structures. This indicates that while LLMs can capture certain spatial relationships, there is considerable room for improvement in their abstract spatial reasoning abilities\cite{yamada2023evaluating}.
\item Function Interpretation: Function interpretation tasks focus on evaluating a system's ability to comprehend functions within the spatial sciences, including their purpose, application contexts, and associated terminology. By replacing specific function names and parameters with abstract identifiers, these tasks emphasize understanding the logic and structure of the functions. This encompasses domains such as coordinate transformation, distance calculation, spherical trigonometry, date processing, multispectral image analysis, and spatial statistical analysis.  
\item Code Explanation: Code explanation tasks assess a system's ability to understand and interpret Python and R code snippets, including the application of specific libraries such as Python's ArcPy or pathlib. By anonymizing function and variable names, these tasks emphasize deep comprehension of code structure and logic rather than reliance on specific naming conventions. By anonymizing function and variable names, these tasks emphasize deep comprehension of code structure and logic rather than reliance on specific naming conventions. As highlighted in the GeoCode-Eval framework proposed by Hou et al., LLMs exhibit potential in code generation and explanation tasks but often suffer from ``code hallucination" or errors due to a lack of domain-specific knowledge and coding datasets\cite{hou2024largelanguagemodelsgenerate}. This suggests that while LLMs can understand and generate code, their performance in complex tasks requires further improvement through domain-specific datasets and optimization strategies.
\item Code Generation: Code generation tasks evaluate a system's ability to generate, adapt, and optimize Python and R code. This includes the application of specific libraries, such as Python's ArcPy and pathlib or R's spatstat and sp, requiring the system to not only execute programming tasks accurately but also demonstrate flexible and innovative use of library functions. Hou et al., in their study of the GeoCode-Eval framework, noted that LLMs show significant potential in code generation tasks, particularly in spatial data processing and geographic modeling. However, due to the lack of domain-specific knowledge and coding datasets, LLMs may encounter challenges such as insufficient accuracy or ``code hallucination" when generating code\cite{hou2024largelanguagemodelsgenerate}. Additionally, Gramacki designed and tested benchmark datasets tailored for geospatial tasks, further demonstrating the strengths and limitations of LLMs in addressing complex geographic problems, while emphasizing the importance of high-quality domain-specific datasets in improving model performance\cite{gramacki2024evaluation}. This indicates that with domain-specific optimizations, LLMs can more effectively support complex programming tasks, particularly in geospatial data science applications.  
\item Code Translation: Code translation tasks evaluate a system's ability to convert code between Python and R. This involves not only basic syntax translation but also the handling of embedded functions from various libraries (e.g., Python’s ArcPy or R’s sp package), spatial reference data, and recursive function conversion. It requires the system to have a deep understanding of the paradigmatic characteristics of both programming languages to ensure the accuracy and practicality of the conversion. 
\end{itemize}

\subsubsection{Spatial Application Tasks}

Spatial application tasks represent the practical implementation of GIS technology in real-world scenarios, serving as a critical link in transforming spatial data, theory, analysis, and operations into tangible value. These tasks directly support societal production and daily life, providing a crucial benchmark for evaluating the spatial application capabilities of LLMs. Based on technical characteristics and application requirements, we categorize spatial application tasks into three core subtasks: NL2API Mapping, Simple Route Planning and toponymic identification.

\begin{itemize}[leftmargin=2em, rightmargin=2em]
\item NL2API Mapping: As noted by Tao and Xu, ChatGPT has demonstrated the ability to generate thematic and mental maps based on text descriptions, providing an innovative solution for mapping tasks. This advancement effectively reduces the barriers in traditional map creation and enhances the efficiency of large-scale mapping. However, the study also emphasizes that current language models remain highly dependent on user intervention when dealing with high-precision details and quality control\cite{tao2023mapping}\cite{xu2024map}. The task is based on the design of the mapping task in the research by Hochmair et al., with the core focus on evaluating the ability of language models to understand user requirements and translate them into specific technical actions, rather than directly generating map images\cite{hochmair2024correctness}. The main challenges of this task are as follows. First, the model must accurately understand the natural language queries posed by users, extract key information, and conduct effective semantic analysis. This requires the model to possess strong language comprehension abilities and to accurately identify geographical locations, directions, and other spatially relevant information within user intent. Second, the model must generate API-compliant requests based on user needs and automatically populate the corresponding parameters. For example, when generating a Mapbox link, the model must not only correctly call the API but also automatically input the appropriate coordinates, style, and zoom level parameters. Finally, the model must translate natural language requirements into specific technical operations, generating executable API requests or programming code. This process tests not only the model's programming capabilities but also its ability to effectively handle API interactions related to mapping, execute spatial data visualization operations, and produce correct technical results. Therefore, this task focuses on assessing the model's ability to translate natural language into specific technical operations, particularly in map generation and geographic information system (GIS) applications. Through this task, we aim to explore the efficient collaboration between language models in processing complex natural language instructions, API interactions, and spatial data visualization.  
\item Simple Route Planning: Simple route planning tasks evaluate a model's ability to perform spatial optimization and intelligent navigation in complex environments. These tasks are typically conducted in grid-based simulated environments, requiring the model to compute optimal paths based on basic directional commands (up, down, left, right) while considering various constraints. As highlighted by Aghzal et al. in their Path Planning from Natural Language (PPNL) benchmark study, LLMs demonstrate some spatial reasoning potential in route planning tasks. Notably, under few-shot prompting conditions, gpt-4 can alternately reason and act, achieving promising results. However, the study also reveals that LLMs show limitations in handling long-term sequential reasoning and planning in more complex environments, particularly when scaled to larger or high-density obstacle scenarios\cite{aghzal2024large}.
\item Toponymic Identification: This task focuses on evaluating a system's ability to extract and comprehend place-name information from complex textual data. The challenge lies in handling non-standardized geographic descriptions from diverse sources such as social media, news reports, and literary works. Systems must demonstrate capabilities in toponym extraction, disambiguation, and contextual understanding, which are critical for the accuracy of geographic information retrieval and location-based services. Research indicates that generative pre-trained models incorporating geographic knowledge (e.g., gpt-4) excel at extracting complex geographic descriptions from disaster-related social media messages, significantly enhancing the recognition of non-standardized geographic descriptions. For instance, using only a small number of geography-informed training samples, these models outperform traditional named entity recognition (NER) tools by over 40\% on disaster datasets\cite{hu2023geo}. Accurate toponymic identification is vital for building high-quality spatial data, particularly in scenarios such as disaster response, where it holds substantial practical value.
\end{itemize}

\subsection{Data Collection}

This study employs a multi-source data integration approach to construct a spatial task dataset, utilizing three main data sources: 1) publicly available datasets, 2) authoritative GIS textbooks, and 3) Wikipedia knowledge snippets. Specifically, for the publicly available datasets, we systematically collected and organized relevant case study datasets used in published GIS-related task studies and academic papers. These datasets include, for example, the 3D trajectory data from the CALVIN benchmark, spatial structure question-answering data from the study by Yamada et al., and spatiotemporal reasoning data from the PPNL benchmark. These datasets provide rich empirical data, covering a variety of spatial task scenarios. Secondly, in reference to authoritative GIS textbooks, we selected classic texts such as Tang's 100 Case Studies in Basic GIS Operations, which provided us with foundational theoretical data and practical spatial task operations. Finally, for the Wikipedia knowledge snippets, we selected entries closely related to the GIS field, covering fundamental geographic concepts and their professional descriptions. After all data collection was completed, we conducted manual screening and organization to ensure the accuracy and professionalism of the knowledge content. 

To ensure the quality of the dataset and the validity of the tasks, we invited six GIS experts to participate in the question design process, including two GIS professors from universities, two PhD students, and two master's students in the GIS field. The expert panel, based on the materials collected from the three aforementioned data sources and considering the specific characteristics of different spatial tasks, designed four categories of questions aimed at comprehensively assessing knowledge understanding and application abilities in spatial task processing.    

\begin{itemize}[leftmargin=2em, rightmargin=2em]
\item The first type is multiple-choice questions, where each question provides four options with only one correct answer. These questions primarily assess geographic literacy.
\item The second type is true/false questions, requiring binary judgments (true/false) on given GIS-related statements. These are also used to evaluate geographic literacy.
\item The third type is objective-answer questions, designed based on deterministic knowledge and requiring precise numerical or textual responses. These questions cover multiple categories, including geographic literacy, toponymic identification, spatial understanding, numerical trajectory recognition, and geographic feature search.
\item The final type is subjective-answer questions, which are highly open-ended and divided into two aspects: theoretical analysis and programming-related tasks. These questions allow for various reasonable answers and are primarily aimed at assessing deeper spatial cognition and practical application abilities. The categories involved include geographic literacy, GIS concepts, NL2API Mapping, code explanation, code generation, code translation, and simple route planning.
\end{itemize}

\subsection{Expert Validation}

Ultimately, we curated and organized a total of 900 questions, covering various aspects from GIS concepts to programming skills (see Figure \ref{figure-2:PNG}). These questions are distributed across 12 distinct categories, including rare and complex questions, aiming to comprehensively cover the knowledge and skills within the GIS domain. To ensure the dataset's high quality, all questions were subject to expert review and revision to guarantee answer accuracy and underwent multidimensional analysis and validation tailored to each question type. For multiple-choice questions, experts focused on verifying the distinctiveness and validity of the options to ensure the uniqueness of the correct answer. For true/false questions, experts established clear and unambiguous evaluation criteria to ensure the rigor and decisiveness of the propositions. For objective-answer questions, experts validated the accuracy of the question descriptions and ensured that the answers were unique and unambiguous. For subjective-answer questions, including procedural and code-related tasks, experts provided thorough evaluations. For text-based questions, experts manually revised them to ensure content completeness, readability, and accuracy. For code-based questions, experts individually validated the executability of the code and the correctness of the results.

\section{Methods}

\subsection{Model and prompt}

In this spatial task dataset test, we included several top-performing LLMs. We selected three models from OpenAI: gpt-3.5-turbo, gpt-4-turbo-2024-04-09, and gpt-4o. The gpt-3.5-turbo is regarded as a milestone in LLMs due to its coherent and human-like responses, making it a benchmark for industry performance evaluations. The gpt-4-turbo-2024-04-09 is widely recognized for its exceptional capabilities, ranking highly on multiple model test leaderboards such as MMLU and MATH. The gpt-4o (``o" stands for ``omni"), OpenAI's latest model, offers enhanced speed and achieves superior performance metrics. We also included ANTHROPIC's Claude-3-Sonnet-20240229, which demonstrates outstanding performance in natural language understanding and generation tasks, representing an optimal balance of performance and processing speed in the Claude series. Additionally, we selected moonshot-v1-8k from Moonshot AI and glm-4 from Zhipuai. The glm-4 exhibits remarkable capabilities in semantic understanding, code generation, and knowledge processing tasks, achieving competitive performance metrics comparable to gpt-4 in several benchmarks. The moonshot-v1-8k similarly demonstrates strong performance across various evaluation metrics and specialized tasks.

Our test is divided into two rounds, aiming to comprehensively evaluate the capabilities of various LLMs in handling spatial tasks based on natural language prompts. The first round is a zero-shot test, which does not involve any prompt tuning to assess the initial performance of the models. This allows us to observe the models' direct response capabilities to spatial tasks. We used the following system prompt for testing: You are a quiz assistant answering the following questions and outputting the results without any additional explanation. 

Following the first round of testing, we will conduct prompt strategy tuning for spatial tasks where performance was suboptimal to evaluate and enhance the LLMs' performance in complex spatial tasks. These strategies include: 

\begin{description}[style=unboxed,leftmargin=0cm]
    \item[\textbf{One-shot:}] In the One-shot test, the model receives an example question and its answer to establish context understanding before immediately facing a new question. This method tests whether the model can quickly learn from a single example and apply it to a new scenario without additional data or extended training. This strategy is particularly useful for evaluating the model's response speed and adaptability when encountering unknown, one-off questions, effectively revealing the model's mechanism for absorbing and processing new information\cite{li-etal-2024-one}. 
    \item[\textbf{Combined Techniques Prompt:}] This strategy uses a combination of model-specific identity prompts, detailed test category descriptions, and specific example questions and answers to create a rich informational environment for the model. This helps the model build a more accurate context for understanding and solving problems and utilize multiple information sources to improve decision accuracy. Combined Techniques Prompt is an advanced testing method aimed at evaluating how models perform complex tasks in an information-rich, structured environment and exploring how different types of information interact to optimize model performance. 
    \item[\textbf{Chain of Thought:}] The Chain of Thought (CoT) method requires the model to receive and process an example problem with a detailed solution process before answering questions. This strategy emphasizes the model's ability to understand and generate logical answers by simulating human problem-solving thinking processes to enhance answer transparency and interpretability. CoT is particularly suitable for handling problems requiring advanced reasoning and logical thinking, demonstrating how the model performs in a continuous logical sequence and systematically processes and solves complex problems\cite{wei2022chain}\cite{wei2022emergent}. 
    \item[\textbf{Zero-shot-CoT:}] Zero-shot-CoT testing involves providing the model with step-by-step thinking instructions (e.g., ``Let's think step by step") without prior specific training or example input, directly guiding the model to develop a thought process. This method's core is to evaluate how the model autonomously constructs a logical chain of problem-solving without direct prior examples. It aims to test the model's autonomous learning and innovation capabilities, particularly its adaptability and solution strategies when facing entirely new and complex problems\cite{kojima2022large}. 
\end{description}

Through these methods, we aim to further explore and understand the potential and limitations of various models in handling specific spatial tasks. 

\subsection{Collecting model answers}

As language models are increasingly applied to handle various complex tasks, ensuring these models provide unbiased and consistent answers has become a significant challenge. To address this need, this study developed an automated script specifically for standardizing the answer collection process, aiming to minimize human error and the potential influence of conversation history through technical means. This script directly calls the APIs of various models, interacting with them using text input only, and adheres to a single-turn dialogue mode. The single-turn dialogue mode was chosen to minimize the possible impact of conversation history on the answers, ensuring that each answer is generated independently of previous dialogue context. This operation mode is particularly suitable for evaluating the model's ability to respond to immediate questions rather than its performance in multi-turn, continuous dialogues. Additionally, to ensure consistency of experimental conditions and to promote more predictable and consistent responses from the models, we uniformly set the temperature parameter to 1 for all models tested. This parameter setting aims to enhance the LLMs' cognitive ability in generating answers, thereby allowing for a more accurate assessment of their true performance levels. 

\begin{figure}
\centering
\includegraphics[width=0.8\linewidth]{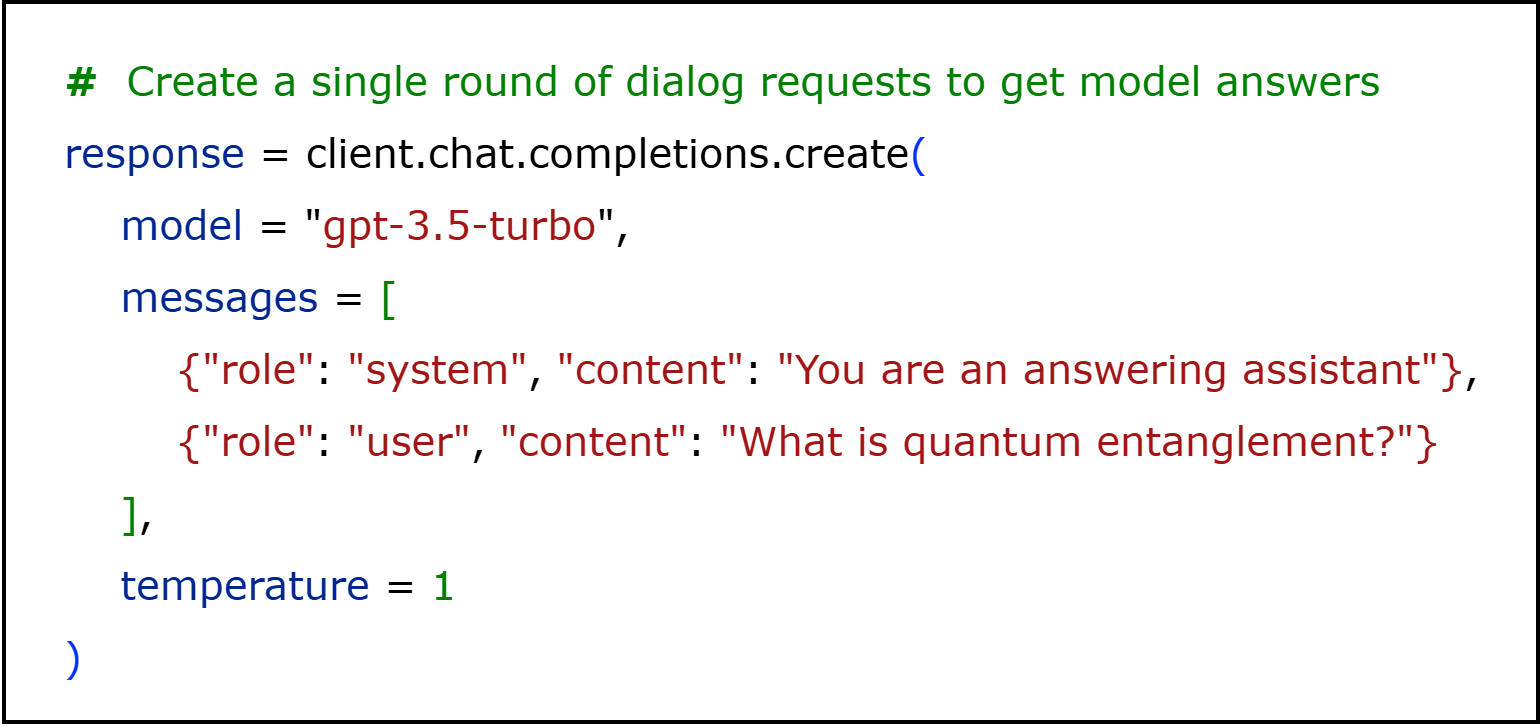}
\caption{\label{figure-3:PNG}An example of conducting a single round of  dialog with the gpt-3.5-turbo model via an API call. }
\end{figure}

Specifically, when testing the gpt-3.5-turbo model, we observed that the model might generate different answers to the same question in different dialogues, particularly for questions with a high degree of subjectivity. This variation is primarily due to the random sampling technique used during decoding. Although these differences are typically minor, to improve the simplicity and efficiency of data processing, we chose to collect only one answer per question. Figure \ref{figure-3:PNG} is a concrete example illustrating the process of conducting a single round of  dialog with the gpt-3.5-turbo model via an API call.

This code example clearly demonstrates how to construct an API request to ensure consistency and standardization of parameters for each request, thereby enhancing the reliability and validity of the study. 

\subsection{Evaluation measures}

Accuracy statistics and qualitative analysis will be used to evaluate the chatbot's responses. In statistical analysis, most answers can be clearly classified as correct or incorrect, such as multiple-choice questions or reasoning questions with clear answers. However, in open-ended tasks like explaining GIS concepts or describing trajectory data visualization, the chatbot's responses may only approximate the correct answer. Such cases, typically found in explanatory or conceptual tasks, will be specially marked with a score of 1. Qualitative analysis will explore the challenges encountered during task execution, along with related examples and illustrations. To accurately assess the model's ability to handle different types of questions, we use the following specific scoring method to quantify the accuracy of the responses:

\begin{itemize}[leftmargin=2em, rightmargin=2em]
\item Fully Correct Answer: When the model provides an answer that is completely correct and meets the expectations, we give it 2 points. This indicates that the model not only understood the core of the question but also generated an accurate and correct answer. 
\item Partially Correct Answer: For answers that are partially correct or partially meet the requirements of the question, we give it 1 point. This situation typically occurs when the model captures part of the key points of the question but does not fully cover all the correct details or mishandles some information. 
\item Completely Incorrect Answer: If the model's answer is unrelated to the correct answer or completely wrong, we give it 0 points. This indicates a fundamental misunderstanding of the question by the model or that the generated response is entirely off the mark. 
\end{itemize}

Human reviewers play a crucial role in the scoring process, especially when dealing with partially correct answers. Certain task categories, such as ``Geographic Literacy" and ``GIS Concepts", often include multiple possible correct answers or require subjective judgment in the responses. In these cases, human judgment is essential for determining the final score. To ensure objectivity and consistency in scoring, we require all reviewers to engage in detailed discussions before scoring and decide the final score through a voting process. For specific task categories involving code execution, such as ``Code Generation" and ``Code Translation", we have made appropriate adjustments to the scoring criteria:

\begin{itemize}[leftmargin=2em, rightmargin=2em]
\item Score of 2: The code runs perfectly, and the output format is correct, fully meeting the expected result. 
\item Score of 1: The code is basically correct but has minor errors or deficiencies, such as syntax errors, logic errors, or format errors. 
\item Score of 0: The code fails to run, or the result deviates significantly from the expected outcome, indicating a clear deficiency in the model's understanding and generation of code. 
\end{itemize}

Through this detailed and systematic scoring mechanism, we can more accurately measure and compare the ability and effectiveness of different models in handling various types of problems. This approach not only helps us identify each model's strengths and weaknesses but also provides valuable data support for future model optimization and development. 

To comprehensively evaluate and compare the ability of LLMs in handling spatial tasks, we introduced Weighted Accuracy (WA) as a key evaluation metric. This metric assigns different weights to the scores of each response to accurately reflect the importance and difficulty of the tasks. The calculation formula for WA is as follows:
\begin{equation}
WA = \displaystyle\frac{2 \cdot n(s2) + 1 \cdot n(s1)}{2 \cdot (n(s0) + n(s1) + n(s2))}
\end{equation}
The variable n(s2) represents the number of answers with a score of 2, n(s1) represents the number of answers with a score of 1, and n(s0) represents the number of answers with a score of 0. 

By applying this formula, we can accurately quantify and compare the efficiency and accuracy of different models in handling spatial-related problems. This provides valuable data support and insights for the optimization and development of future models. This approach emphasizes the importance of a comprehensive assessment of model performance, ensuring that our evaluation is as thorough and impartial as possible. 

\section{Results and Analysis}

After the first round of zero-shot testing, we evaluated the total scores and WA of 12 subspace tasks, as shown in the table~\ref{tab:t1}. In this evaluation, gpt-4o performed exceptionally well, achieving the highest number of scores of 2 and the highest overall WA. Following closely is gpt-4-turbo-2024-04-09, whose scores are almost identical to gpt-4o. Although Claude-3-sonnet-20240229 exceeded glm-4 in the number of questions scored 2, glm-4's overall WA remained slightly higher. The overall slightly lower performers are moonshot-v1-8k and gpt-3.5-turbo. In the evaluation of 900 spatial tasks, gpt-4o provided completely correct answers for 627 questions and partially correct answers for 24 questions, with an overall WA of 71\%. The gpt-4-turbo-2024-04-09 had an overall WA of 69.7\%, just 1.3\% lower than gpt-4o. Comparatively, gpt-3.5-turbo performed the worst, with an overall WA of only 43.8\%. As the most advanced model released by Zhipu AI, glm-4's performance ranked after gpt-4o and gpt-4-turbo-2024-04-09, with an overall WA of 62.4\%. The claude-3-sonnet-20240229 and moonshot-v1-8k had overall weighted accuracies of 62.1\% and 53.2\%, respectively. Analyzing the score distribution, we observed that the number of questions scored 1 was significantly lower than those scored 0 and 2, indicating that the models were able to provide clear answers in most cases. It is noteworthy that the performance gap between gpt-4o and gpt-3.5-turbo is substantial, which aligns with the conclusions of gpt-4o's latest report\cite{openai2024gpt4o}. 

\begin{table}[!htb]
\centering
\caption{\label{tab:t1}Scores of each model and overall WA.}
\begin{tabular}{@{}lcccccc@{}}
\toprule
\textbf{Model} & \textbf{Count S0} & \textbf{Count S1} & \textbf{Count S2} & \textbf{WA (\%)} \\
\midrule
gpt-3.5-turbo & 481 & 50 & 369 & 43.8 \\
gpt-4o & 249 & 24 & \textbf{627} & \textbf{71.3} \\
gpt-4-turbo-2024-04-09 & 260 & 26 & 614 & 69.7 \\
claude-3-sonnet-20240229 & 334 & 15 & 551 & 62.1 \\
moonshot-v1-8k & 409 & 24 & 467 & 53.2 \\
glm-4 & 318 & 41 & 541 & 62.4 \\
\bottomrule
\end{tabular}
\end{table}

\begin{figure}[!htb]
\centering
\includegraphics[width=1\linewidth]{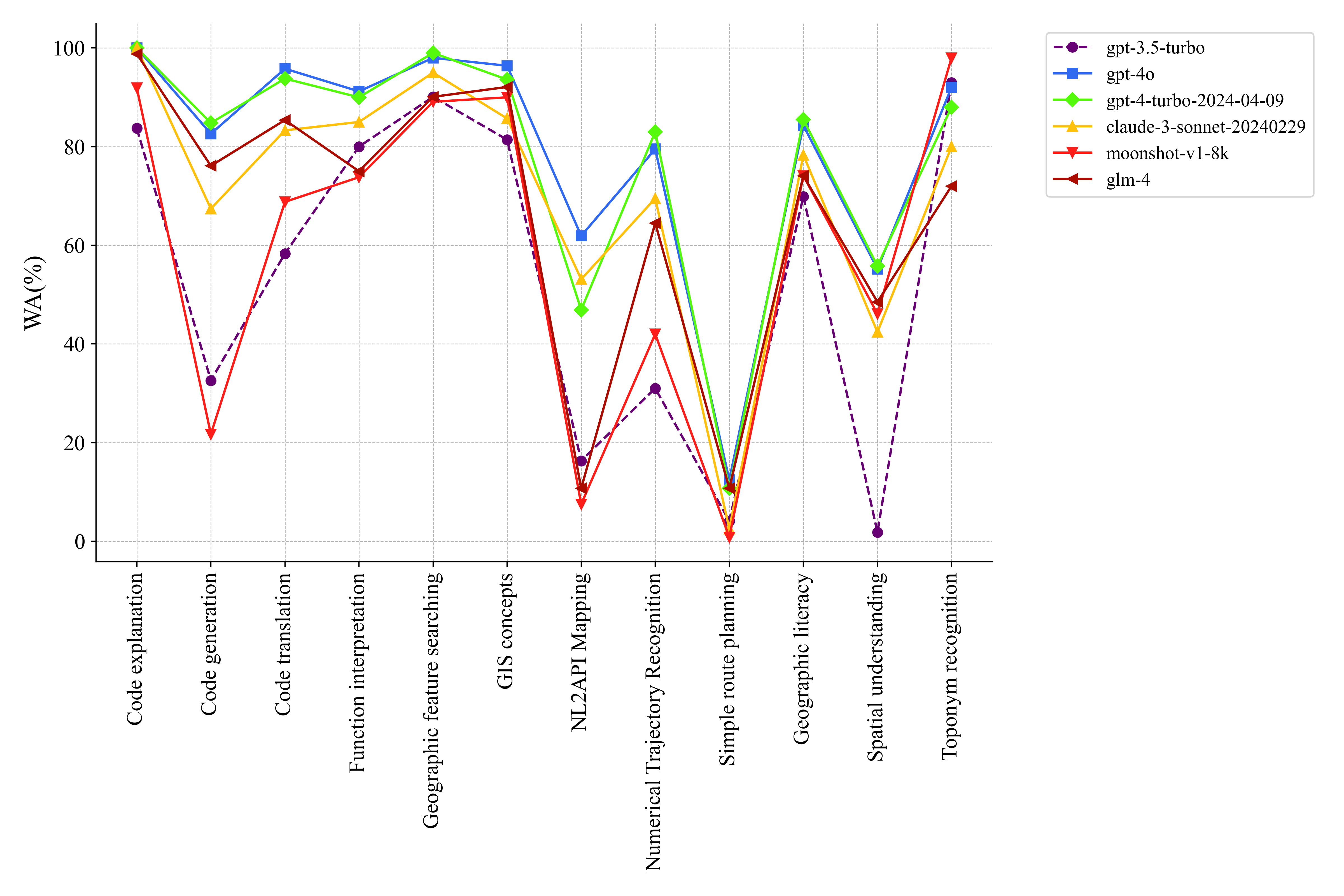}
\caption{\label{figure-4:PNG}Comparison of model results in zero-shot testing. }
\end{figure}

Figure \ref{figure-4:PNG} shows the overall WA of various models across different spatial sub-tasks. The figure indicates that the performance order of the models remains consistent across most spatial sub-tasks. Additionally, Table~\ref{tab:t2} provides a detailed breakdown of the results for each spatial sub-task. The gpt-4o and gpt-4-turbo-2024-04-09 performed the best in six different categories and achieved 100\% accuracy in code explanation tasks. In contrast, gpt-3.5-turbo ranked last in performance across six categories. 

\subsection{Model accuracy}

The overall performance of the models on explanatory, knowledge-based, and conceptual tasks was quite good, especially in GIS conceptual tasks, where all models achieved an overall accuracy (WA) of over 80.0\%. This performance demonstrates that the training corpora of LLMs encompass extensive GIS knowledge and showcase the models' strong capability in understanding and applying this knowledge. However, in numerical trajectory recognition tasks, gpt-3.5-turbo had an accuracy of only 31.0\%, significantly lower than the other models. This discrepancy may be related to its training dataset, parameter settings, or training algorithm. Additionally, when it comes to operational tasks involving specific programming languages and code details, gpt-3.5-turbo and other weaker models may underperform due to insufficient specialized training. 

\begin{table}[htbp]
\centering
\caption{\label{tab:t2}Performance metrics by category and model.\\(G3.5t:gpt-3.5-turbo,G4o:gpt-4o,G4t:gpt-4-turbo-2024-0409,Cs:claude-3-sonnet-20240229,Ms:moonshot-v1-8k,Glm:glm-4)}
\begin{tabular}{@{}lcccccc@{}}
\toprule
\multirow{2}{*}{\textbf{Category}} & \multicolumn{6}{c}{\textbf{WA (\%)}} \\ \cmidrule(lr){2-7}
 & \textbf{G3.5t} & \textbf{G4o} & \textbf{G4t} & \textbf{Cs} & \textbf{Ms} & \textbf{Glm} \\ \midrule
Code explanation & 83.7 & \textbf{100.0} & \textbf{100.0} & \textbf{100.0} & 91.9 & 98.8 \\
Code generation & 32.6 & 82.6 & 84.8 & 67.4 & 21.7 & 76.1 \\
Code translation & 58.3 & \textbf{95.8} & 93.8 & 83.0 & 68.8 & 85.4 \\
Function interpretation & 80.0 & 91.2 & \textbf{95.2} & 88.5 & 73.8 & 75.0 \\
Geographic feature searching & 93.1 & \textbf{98.1} & 96.4 & 96.0 & 90.0 & 92.1 \\
GIS concepts & 81.0 & 96.4 & 93.6 & 96.0 & 94.9 & 92.1 \\
NL2API Mapping & 16.3 & \textbf{61.9} & 46.9 & 53.1 & 7.5 & 38.8 \\
Numerical Trajectory Recognition & 31.0 & 79.5 & \textbf{83.0} & 69.0 & 42.0 & 64.5 \\
Simple route planning & 4.1 & \textbf{12.4} & 8.7 & 2.5 & 0.0 & 7.5 \\
Geographic literacy & 69.9 & 85.0 & 85.5 & 84.0 & 71.7 & 80.6 \\
Spatial understanding & 1.8 & \textbf{55.2} & 40.8 & 54.8 & 46.1 & 45.0 \\
Toponym recognition & \textbf{93.0} & 92.0 & 88.0 & 88.0 & \textbf{98.0} & 72.0 \\
\bottomrule
\end{tabular}
\end{table}

The models generally performed poorly on reasoning and application tasks. Especially in simple route planning tasks, which require the models to not only understand the specific requirements but also perform complex logical reasoning, such as considering multiple possible routes and evaluating various environmental factors. Even the best-performing model, gpt-4o, had a WA of only 12.4\%, while the worst-performing model, moonshot-v1-8k, had a WA of just 0.8\%. Additionally, in spatial understanding tasks, gpt-3.5-turbo had a WA of only 1.8\%, significantly lower than the other models. In NL2API Mapping tasks, which require models to generate Mapbox links for specified areas or perform map visualization based on specific commands, the performance gap between the models was significant. The gpt-4o performed the best with a WA of 61.9\%, while moonshot-v1-8k performed the worst with a WA of only 7.5\%, 54.4\% lower than gpt-4o. 

Overall, although the major models performed well in explanatory, knowledge-based, and conceptual tasks, particularly demonstrating excellent accuracy in GIS conceptual tasks, their performance in reasoning and application tasks was generally poor. simple route planning and NL2API Mapping tasks especially highlighted the challenges models face in complex logical reasoning and handling advanced spatial information. These results indicate that while current large models can achieve high accuracy in certain areas, they still require further optimization and training for tasks that demand high-level reasoning and specialized knowledge. 

\subsection{Clustering questions by difficulty}

To conduct a more detailed model evaluation, we further divided the spatial task dataset into three levels based on difficulty: Level I (easy), Level II (medium), and Level III (difficult). Through in-depth discussions with an expert panel, we established specific criteria for classifying the difficulty of the questions: 1) Questions correctly answered and scored 2 by at least five models were classified as easy. 2) Similarly, questions correctly answered and scored 2 by at least two but no more than four models were marked as medium difficulty. 3) Questions correctly answered and scored 2 by at most one model (usually the best-performing model, such as gpt-4o) were defined as difficult, as these questions are often the most challenging. In these three difficulty levels, the number of questions was 395 (easy), 275 (medium), and 230 (difficult), totaling 900 questions. Next, we will evaluate the performance of each model based on these difficulty levels (see Table~\ref{tab:t3} for the results). The evaluation criteria are as follows:

\begin{itemize}[leftmargin=2em, rightmargin=2em]
\item If the model answers an easy question completely correctly (score of 2), it is considered to have correctly solved one easy question. 
\item If the model answers an easy question partially correctly (score of 1), it is considered to have correctly solved 0.5 of an easy question. 
\item The evaluation criteria for medium and difficult questions are the same. 
\end{itemize}

\begin{table}[!htb]
\centering
\caption{\label{tab:t3}The model's scores on questions of different difficulty levels.}
\begin{tabular}{lccc}
\toprule
\textbf{Model} & \textbf{Easy} & \textbf{Medium} & \textbf{Difficult} \\ \midrule
gpt-3.5-turbo & 342 & 96.5 & 7.5 \\
gpt-4o & 385.5 & 200.5 & \textbf{29} \\
gpt-4-turbo-2024-04-09 & \textbf{391} & \textbf{207} & \textbf{29} \\
claude-3-sonnet-20240229 & 388 & 160 & 10.5 \\
moonshot-v1-8k & 375 & 95.5 & 8.5 \\ 
glm-4 & 368 & 141 & 25 \\
\bottomrule
\end{tabular}
\end{table}

\begin{figure}[!htb]
\centering
\includegraphics[width=0.8\linewidth]{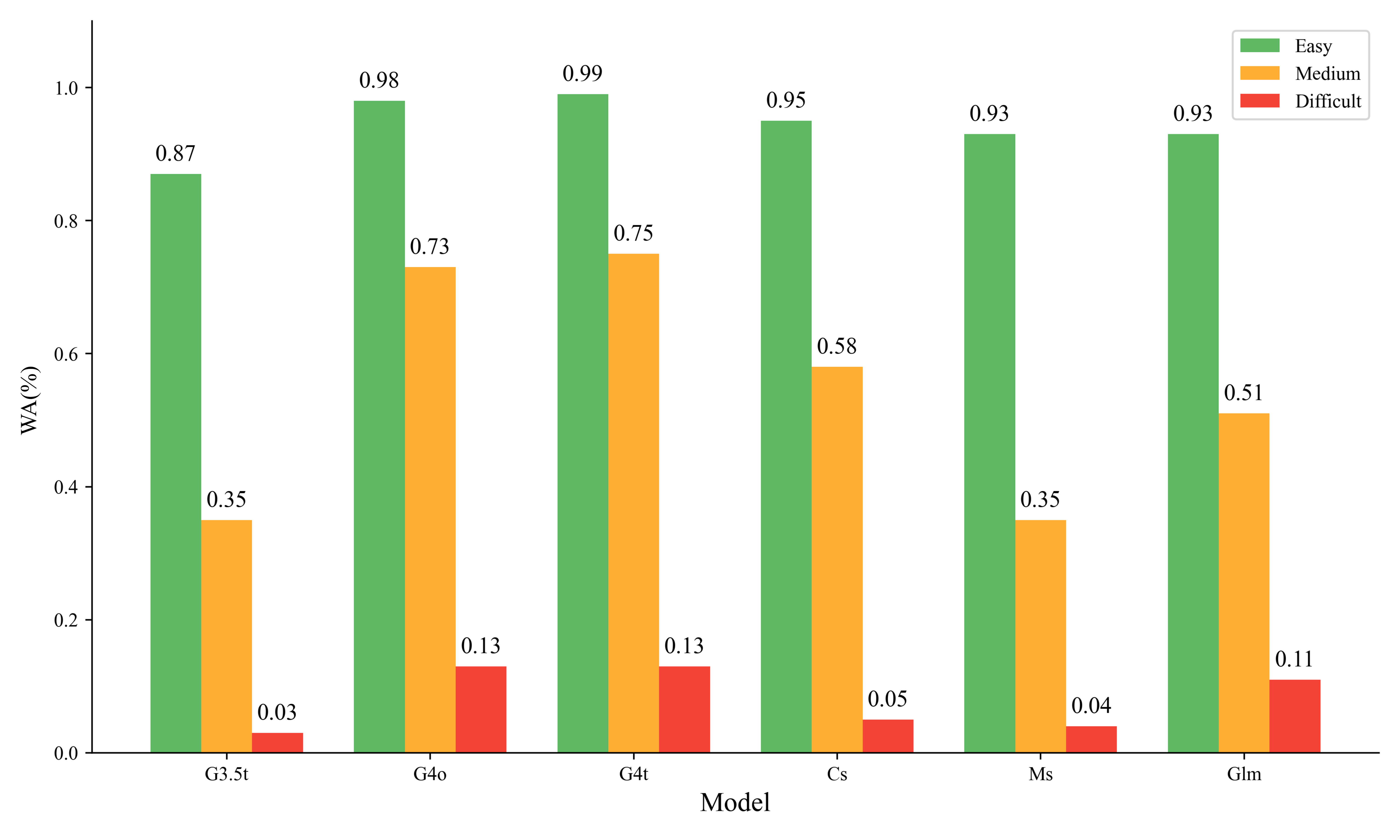}
\caption{\label{figure-5:PNG}Accuracy of questions clustered according to difficulty. (G3.5t:gpt-3.5-turbo,G4o:gpt-4o,G4t:gpt-4-turbo-2024-0409,Cs:claude-3-sonnet-20240229,Ms:moonshot-v1-8k,Glm:glm-4)}
\end{figure}

According to Figure \ref{figure-5:PNG}, as expected, the models generally performed well on easy questions. The gpt-4-turbo-2024-0409 performed the best with an accuracy rate of 0.99, while gpt-4o and Claude-3-sonnet-20240229 also had high accuracy rates of 0.98. On medium difficulty questions, gpt-4-turbo-2024-0409 also led with an accuracy rate of 0.73. The gpt-3.5-turbo and moonshot-v1-8k performed poorly, with an accuracy rate of only 0.35. In handling difficult questions, all models performed generally poorly. The gpt-3.5-turbo stood out slightly with an accuracy rate of 0.03, while gpt-4o and gpt-4-turbo-2024-0409 had similar performances, both at 0.13. This indicates that even the latest models faced challenges with the hardest questions. Overall, gpt-4-turbo-2024-0409 and gpt-4o performed well across all three difficulty levels, while gpt-3.5-turbo performed poorly. 

It is noteworthy that using the models' answers to classify the difficulty of questions is a simple and efficient method. Additionally, another possible approach is to seek the opinions of an expert panel to determine the difficulty level of each question. The main purpose of classifying the dataset by difficulty is to evaluate the performance of future models on this dataset, monitor their progress over time, and compare it with human performance. For example, if a model only shows high accuracy on easy questions, we cannot consider it a powerful model based on this alone. Conversely, if the model also performs well on medium and difficult questions, we can consider it an outstanding and effective model for handling spatial tasks. Therefore, the ability to handle questions of varying difficulty is crucial when evaluating a model's capabilities. 

\subsection{Prompt tuning}

After conducting the first round of zero-shot testing, we found that the models performed poorly in simple route planning, spatial understanding, and NL2API Mapping tasks. Additionally, based on the previous difficulty classification results, we observed that the number of difficult questions in these three tasks was relatively high. Therefore, we will conduct specific prompt optimization tests for these three tasks. 
For the simple route planning and spatial understanding tasks, we used four prompt strategies for optimization testing: One-shot, Combined Techniques Prompt, Chain of Thought, and Zero-shot-CoT. When discussing the NL2API Mapping task, considering that code-related issues are not suitable for prompt strategy optimization, we focused only on the first type of NL2API Mapping task (see Table~\ref{tab:t4}): where the model accesses online map services to construct and return a map link for a specific area. Since this type of NL2API Mapping task is not suitable for testing with reasoning-based prompt strategies, we selected One-shot and Combined Techniques Prompt strategies for optimization testing. For the simple route planning task, examples of system prompts for each prompt strategy are shown below. These prompt examples are provided as system prompts under the ``system" role.

\begin{enumerate}[label=\textbf{(\alph*)}, align=left, itemsep=4pt] 
    \item \textbf{One-Shot} \normalsize\vspace{6pt}\\
    Question: You are in a 4 by 4 two-dimensional array. You have to avoid some obstacles: (0,3). From (1,1) to (3,0). Just use ``up, down, left, right" to indicate your route. \\
    Answer: path: down, down, left

    \item \textbf{Combined Techniques Prompt}\normalsize\vspace{6pt}\\
    As a simple route planning expert with extensive experience in GIS, you will be responsible for designing and analyzing optimal paths that avoid obstacles. \\
In the ``simple route planning" category, the task is to navigate from a starting point to an end point while avoiding obstacles placed in the environment. The process involves computing an efficient route, usually in a gridded virtual environment, by using simple directional commands such as ``up", ``down", ``left", ``right". \\
    Question: You are in a 4 by 4 two-dimensional arrays. You have to avoid some obstacles: (0,3). From (1,1) to (3,0). Just use ``up, down, left, right" to indicate your route. Your output is formatted as: path:\\
    Answer: path:down,down,left

    \item \textbf{Chain of Thought (CoT)}\normalsize\vspace{6pt}\\
    Question: You are in a 4 by 4 two-dimensional arrays. You have to avoid some obstacles: (0,3). From (1,1) to (3,0). Just use ``up, down, left, right" to indicate your route. Your output is formatted as: path:\\
    Answer: From the starting point of (1,1), you need to reach the end of (3,0). In the process of moving, we need to avoid the obstacle located at (0,3). First move down one step to (2,1), then continue down one step to (3,1), and finally move one step to the left to reach (3,0) to complete the path.  So the answer is: path: down, down, left. 

    \item \textbf{Zero-shot-CoT}\normalsize\vspace{6pt}\\
Let’s think step by step. 
\end{enumerate}

\begin{table}[ht]
\centering
\caption{\label{tab:t4}Results of the first type of NL2API Mapping task.}
\begin{tabular}{@{}lcccc@{}} 
\toprule
\textbf{Model} & \textbf{Count S0} & \textbf{Count S1} & \textbf{Count S2} & \textbf{WA (\%)} \\ 
\midrule
gpt-3.5-turbo & 46 & 0 & 13 & 22.0 \\
gpt-4o & 19 & 0 & 40 & \textbf{67.8} \\
gpt-4-turbo-2024-04-09 & 27 & 0 & 32 & 54.2 \\
claude-3-sonnet-20240229 & 28 & 0 & 31 & 52.5 \\
moonshot-v1-8k & 53 & 0 & 6 & 10.1 \\
glm-4 & 37 & 0 & 22 & 37.2 \\
\bottomrule
\end{tabular}
\end{table}

\begin{description}[style=unboxed,leftmargin=0cm]
    \item[\textbf{Simple route Planning Task:}] By comparing the test results (see Figure \ref{figure-6:PNG}), we observed that optimizing input prompts significantly improved the models' overall accuracy. For gpt-4o, in the initial unoptimized zero-shot test, the model's WA was only 12.4\%. After the second round of prompt optimization, its WA increased to 50.8\%, 43.3\%, 87.5\%, and 68.3\% under the One-shot, Combined Techniques Prompt, CoT, and Zero-shot-CoT strategies, respectively. We observed that the CoT and Zero-shot-CoT strategies had a particularly significant effect on improving the model's WA. For gpt-4-turbo-2024-04-09, similar improvement trends were observed in the corresponding tests. The gpt-3.5-turbo achieved its highest WA of 25.8\% under the Zero-shot-CoT strategy but showed a performance decline in the One-shot test. Similarly, moonshot-v1-8k's WA dropped to 0 in the One-shot test, and it did not exceed 10\% under the CoT and Zero-shot-CoT strategies. The claude-3-sonnet-20240229 showed no significant improvement with the Combined Techniques Prompt and Zero-shot-CoT strategies, achieving its best performance of 9.1\% with the CoT strategy. For glm-4, it achieved its best performance of 36.6\% with the CoT strategy. Notably, its WA decreased in both the One-shot and Combined Techniques Prompt tests, dropping to 0 in the Combined Techniques Prompt test. 
\begin{figure}[!htb]
\centering
\includegraphics[width=0.9\linewidth]{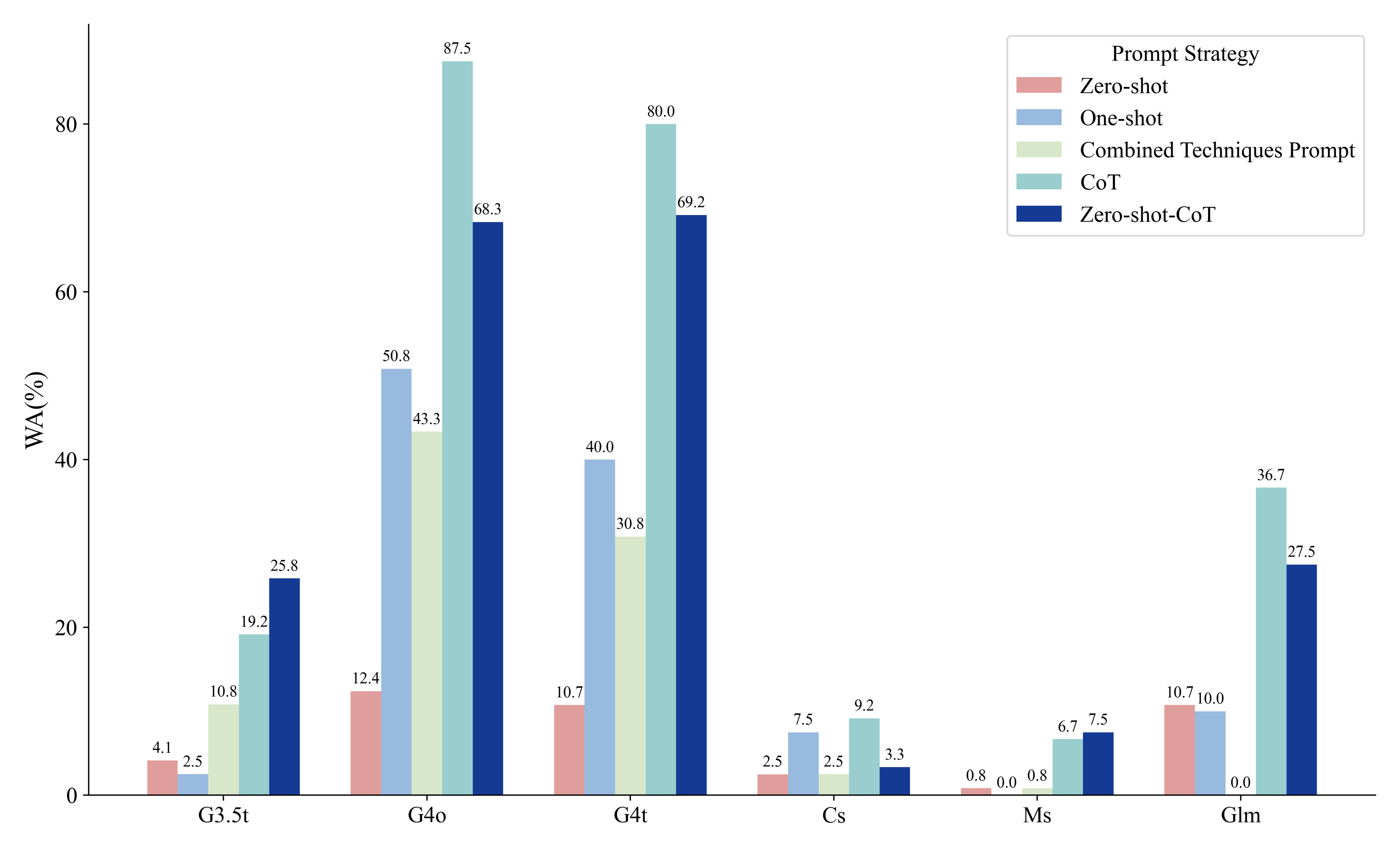}
\caption{\label{figure-6:PNG} Prompt optimization test results for simple route planning task. (G3.5t:gpt-3.5-turbo, G4o:gpt-4o, G4t:gpt-4-turbo-2024-0409, Cs:claude-3-sonnet-20240229, Ms:moonshot-v1-8k, Glm:glm-4)}
\end{figure}
    \item[\textbf{Spatial Understanding Task:}] After prompt optimization (see Figure \ref{figure-A1:PNG}), different prompt strategies significantly improved the WA of gpt-3.5-turbo, with the CoT strategy being the most effective, reaching a WA of 37.8\%, which is a 36\% increase from the zero-shot test's 1.8\%. The gpt-4o and gpt-4-turbo-2024-04-09 showed similar growth trends, both achieving their highest WA under the Zero-shot-CoT strategy. For Claude-3-sonnet-20240229, the prompt strategies did not significantly improve its WA, with the CoT strategy being the most effective at 51.2\%, an 8.8\% increase from the zero-shot test's 42.4\%. Additionally, it is worth noting that both Moonshot and GLM models performed poorly in this task after prompt optimization, with different prompt strategies even leading to a decline in performance. The moonshot-v1-8k only showed improvement under the Combined Techniques Prompt strategy, increasing from 46.1\% in the zero-shot test to 50.6\%. The glm-4's performance decreased across all prompt strategies, with the lowest WA of 32.9\% under the Zero-shot-CoT strategy, a 15.6\% drop from the zero-shot test's 48.5\%. 
    \item[\textbf{NL2API Mapping Task:}] By comparing the test results (see Figure \ref{figure-A2:PNG}), we found that the gpt-4o, gpt-4-turbo-2024-04-09, and Claude-3-sonnet-20240229 models showed similar growth trends under the One-shot and Combined Techniques Prompt strategies. The highest weighted accuracies (WA) for these three models were 84.7\%, 76.3\%, and 66.1\%, respectively. Notably, the Combined Techniques Prompt strategy significantly improved the gpt-3.5-turbo model, with its WA increasing from 22.0\% in the zero-shot test to 62.7\%. The One-shot strategy also brought substantial improvement to the moonshot-v1-8k model, with its WA soaring from 10.1\% in the zero-shot test to 76.3\%, matching the performance of gpt-4-turbo-2024-04-09. However, the glm-4 model performed relatively poorly, with its performance even declining under the One-shot strategy. In the Combined Techniques Prompt strategy, its performance only slightly increased from 37.2\% in the zero-shot test to 40.7\%. 
\end{description}

\subsection{Example analysis}

\begin{figure}[!htb]
\centering
\subfigure[Results of gpt-4o.]{
\resizebox*{11cm}{!}{\includegraphics{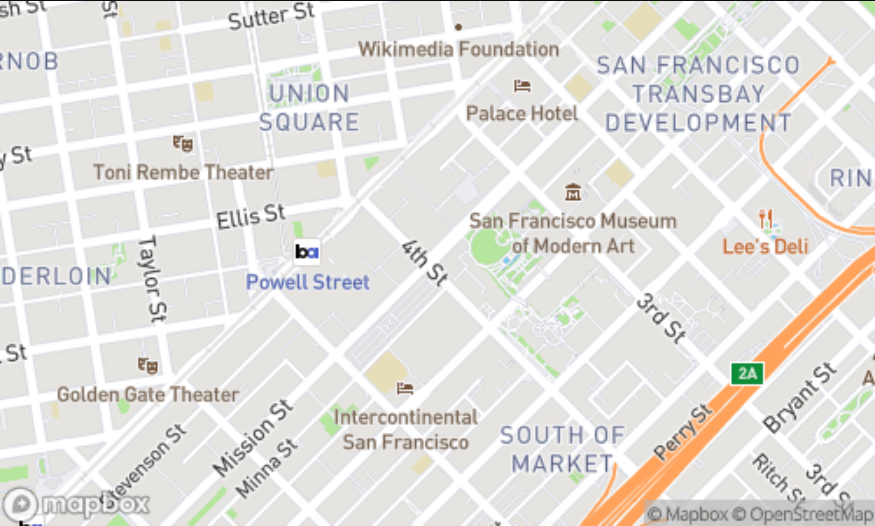}}}\\
\subfigure[Results of gpt-4-turbo-2024-04-09.]{
\resizebox*{11cm}{!}{\includegraphics{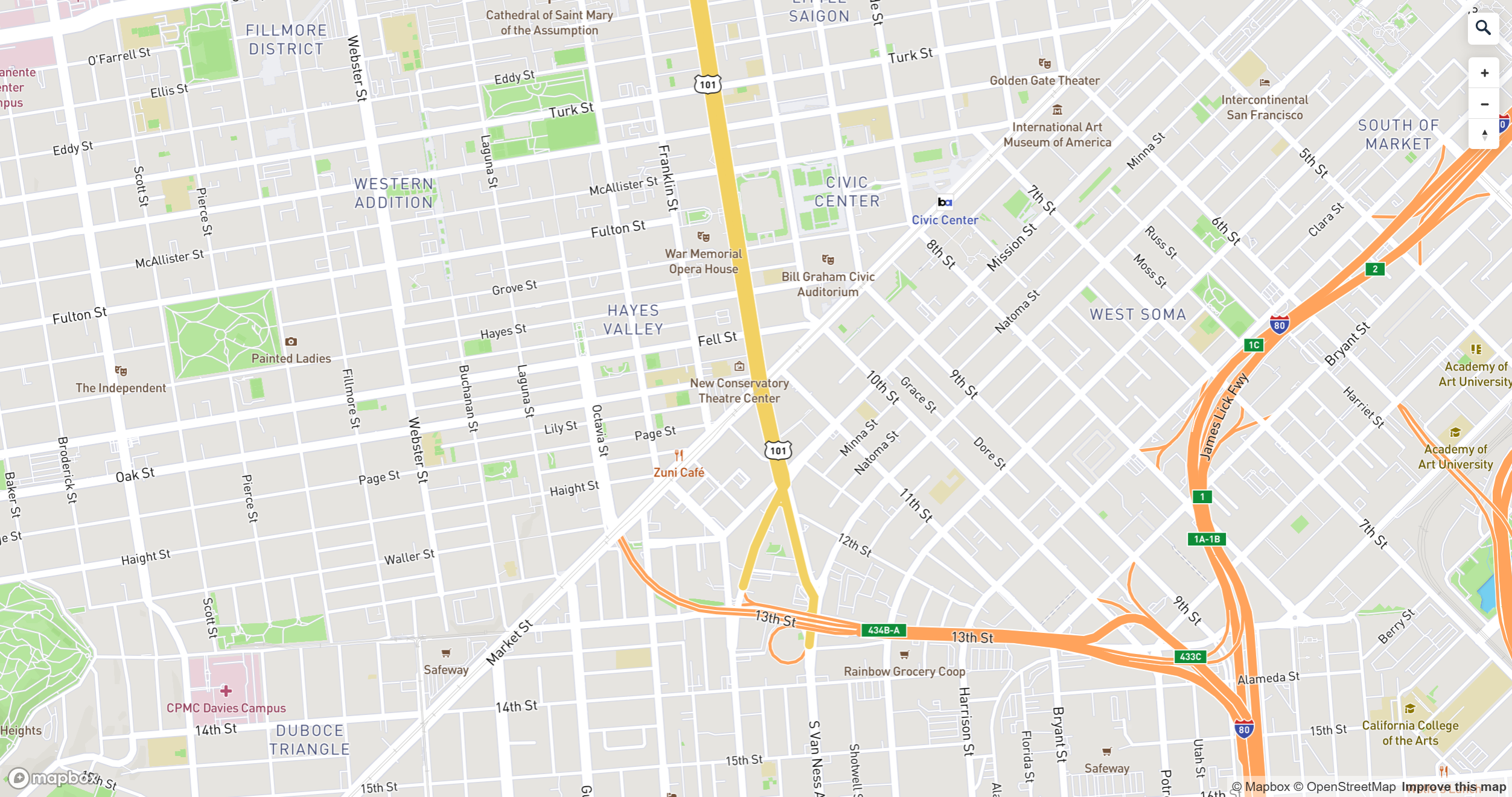}}}
\caption{Results of ``Create a Mapbox map link showing a standard map view of downtown San Francisco" using gpt-4o (top) and gpt-4-turbo-2024-04-09 (bottom).} 
\label{figure-7}
\end{figure}

When tasked with ``creating a Mapbox map link displaying the standard view of downtown San Francisco", we found that both gpt-4o and gpt-4-turbo-2024-04-09 accurately captured the semantic essence of the task and successfully returned a Mapbox link showing downtown San Francisco. However, each model displayed significant differences in their semantic understanding when determining the specific location of ``downtown San Francisco". This difference was not only reflected in the selection of the map's central point but also highlighted the models' differing interpretations of geographic concepts and urban space (see Figure \ref{figure-7}). Specifically, gpt-4-turbo-2024-04-09 selected a central point near the Herbst Theatre, located in the cultural and governmental area of downtown, close to the famous Civic Center. In contrast, gpt-4o centered the map on the San Francisco Museum of Modern Art (SFMOMA), a location closer to the Financial District and Union Square, representing a downtown area with a more commercial and artistic atmosphere. Although both models received perfect scores on the task, they exhibited subtle yet interesting differences in their understanding of ``downtown". This difference highlights the complexity of LLMs in handling geographic and spatial concepts: different models, based on their training data and internal representations, may generate slightly different spatial interpretations of the same geographic concept. This finding not only demonstrates the fine-grained nature of semantic understanding in models but also serves as a reminder that when evaluating language model performance, attention must be given to their ability to handle complex, ambiguous concepts with nuance.

It is important to note that these differences are not due to the models directly manipulating the map's style or content, but rather stem from the models' inherent mechanisms for translating the semantic concept of ``downtown" into specific geographic coordinates. The final presentation of the map is primarily controlled by Mapbox's style definitions and zoom levels; however, the model's ability to select API parameters still plays a critical role in the specific map display. This case vividly illustrates the unique ability of language models to bridge the gap between spatial concept understanding and technical implementation. Specifically, while Mapbox's vector tile system and predefined styles determine the visual details of the map, the model plays a crucial role in translating natural language requirements into precise technical parameters, such as coordinates and zoom levels. The differences in semantic understanding of ``downtown" across models are directly reflected in the selected map center and API parameters, which in turn affect the final map presentation. These subtle differences reveal the complexity of LLMs in processing spatial concepts and their unique potential in bridging the gap between semantic understanding and technical implementation.

In the first round of zero-shot testing, we observed an interesting phenomenon in the Numerical Trajectory Recognition task. When handling the question ``What is the directional description of the coordinate sequence [(5, 0), (5, 15), (15, 15), (15, 5), (5, 5)] in order from left to right?" gpt-3.5-turbo showed a different understanding compared to other models. Specifically, it considered the move from (5, 5) to (5, 0) as the last direction change, thus generating five directional descriptions for the question. To further investigate whether this phenomenon was common, we selected five different cases and conducted three rounds of testing on gpt-3.5-turbo (see Table~\ref{tab:t5}). In these tests, we focused on the number of directional descriptions rather than their accuracy. The results showed that in 15 tests, only 5 successfully generated the correct number of descriptions. 

\begin{table}[htbp]
\centering
\caption{\label{tab:t5}Test results of gpt-3.5-turbo on different cases. The bolded content indicates the number of direction descriptions is accurate.}
\resizebox{\textwidth}{!}{%
\begin{tabular}{p{6cm} p{4cm} p{4cm} p{4cm}}
\toprule
\textbf{Question} & \textbf{Model Answer 1} & \textbf{Model Answer 2} & \textbf{Model Answer 3} \\ \midrule
What is the direction description of the coordinate sequence \texttt{[(10, 10), (10, 20), (20, 20), (10, 20)]} in order from left to right? & north, east, south, west & right, up, right, down & east, north, east, south \\ \midrule
What is the directional description of the coordinate sequence \texttt{[(0, 0), (0, 10), (10, 10), (10, 0)]} in order from left to right? & \textbf{north, east, south} & east, north, west, south & north, east, south, west \\ \midrule
What is the directional description of the coordinate sequence \texttt{[(90, 100), (110, 100), (110, 110), (100, 110), (100, 100)]}in order from left to right? & right, up, right, up, left & \textbf{right, up, left, down} & east, north, east, south, west \\ \midrule
What is the directional description of the coordinate sequence \texttt{[(15, 15), (15, 25), (25, 25), (25, 15)], (10, 15)]} in order from left to right? & east, north, west, south, east & \textbf{north, east, south, west} & \textbf{up, right, down, left} \\ \midrule
What is the directional description of the coordinate sequence \texttt{[(50, 50), (60, 50), (60, 60), (70, 60), (80, 60), (80, 50), (90, 50), (90, 40), (100, 40), (110, 40), (110, 50)]} in order from left to right? & east, east, north, east, east, south, east, south, east, east, north & east, south, east, north, east, south, east, south, east, north, east & \textbf{east, north, east, east, south, east, south, east, east, north} \\ 
\bottomrule
\end{tabular}%
}
\end{table}

\section{Discussion}

\subsection{Performance Comparison of LLMs in Spatial Tasks}

Our research indicates that the gpt-4o model outperforms other models overall, particularly excelling in complex reasoning tasks, with the gpt-4-turbo-2024-04-09 model following closely behind. However, in certain specific sub-tasks, the moonshot-v1-8k model even surpasses both gpt-4o and gpt-4-turbo-2024-04-09 in toponymic identification tasks. For instance, when asked the question ``The Tokyo Tower was badly damaged in last night's earthquake. Which parts of the sentence represent location descriptions?", moonshot-v1-8k correctly identified ``Tokyo Tower" as the location description, whereas gpt-4o mistakenly identified both ``Tokyo Tower" and ``last night" as location descriptions. This suggests that the performance of moonshot-v1-8k in semantic recognition and understanding tasks is on par with that of gpt-4o.

Hu et al. reported that gpt-4 and its variant (Geo-gpt-4) achieved approximately 15\% higher accuracy than gpt-3 and its variants in disaster-related toponymic identification tasks, establishing them as the best-performing models\cite{hu2023geo}. In our study, we observed that moonshot-v1-8k outperformed gpt-4o in toponymic identification tasks, further highlighting the performance variations among different models for such tasks. Nevertheless, gpt-4o demonstrated outstanding performance in handling complex tasks, consistent with our expectations for the model. Compared to the findings of Yamada et al. on spatial understanding capabilities of LLMs, their study revealed that gpt-4 improved accuracy by over 50\% compared to gpt-3.5-turbo in handling complex spatial understanding tasks\cite{yamada2023evaluating}. Our results indicate that gpt-4o's performance in complex reasoning tasks aligns with the conclusions of Yamada et al.'s study, particularly in tasks involving multiple spatial relationships, where gpt-4o exhibited exceptional reasoning capabilities. Additionally, Sharma et al., in their study on spatial reasoning abilities, highlighted that gpt-4 outperformed gpt-3.5 by over 30\% on the CALVIN dataset and achieved a 20\% improvement in accuracy using few-shot prompting strategies\cite{sharma2023exploring}. This finding is consistent with our research, further confirming gpt-4's advantage in spatial reasoning tasks. We also noted that Liu et al., in their study on geographic diversity, observed regional variations in model performance on geographic diversity tasks, particularly in geographic guessing related to UNESCO World Heritage sites\cite{liu2024measuring}. Building on this, we optimized the experiment by employing various prompting strategies to evaluate the performance of each model. The results indicated that the gpt-4 series continued to perform best, with other models showing some improvements as well. Lastly, Aghzal et al., in their study on the path-planning capabilities of LLMs, found that model performance declined significantly as task complexity and grid size increased\cite{aghzal2024large}. Our findings align with this observation. For simple route planning tasks, gpt-4o provided relatively accurate solutions, while gpt-3.5-turbo and moonshot-v1-8k performed poorly, offering incorrect path-planning proposals. In more complex environments, such as route planning in grids with multiple obstacles, gpt-4o was able to provide accurate solutions for most tasks, whereas gpt-3.5-turbo achieved an accuracy rate of only 4\%.

LLMs, as powerful AI tools, exhibit remarkable capabilities but also have their respective strengths and limitations. Therefore, in practical applications, the choice of the appropriate model should be based on specific requirements and contexts. For instance, for simple route planning tasks, the gpt-4o model, combined with a Chain-of-Thought (CoT) prompting strategy, may yield the most effective results. This underscores the practical significance and application value of our research.

\subsection{The Role of Dataset Difficulty Classification}

In traditional computational studies, such as Borji's research on large model evaluation and dataset creation, the discussion on dataset categorization divides the dataset into simple, medium, and difficult categories, where easy questions are those that all models can answer correctly, and difficult questions are those that even the best models fail to answer correctly\cite{borji2023battle}. Building on this framework, we introduced an improvement by using the performance of six LLMs in zero-shot testing to determine the difficulty levels more effectively. In this study, we adopted an innovative strategy based on the performance of models in the first round of zero-shot testing to classify the dataset difficulty. This method integrates the test results of six different LLMs to categorize the difficulty levels of the dataset and further discuss the outcomes.

We believe this strategy is not only more efficient but also significantly promotes subsequent research. Difficulty classification is extremely beneficial for follow-up research, as it allows researchers to more accurately assess the effectiveness of different models in handling simple, medium, and difficult problems. Through this approach, researchers can deeply analyze the models' capabilities and limitations in complex situations. Difficulty classification guides further model development, especially when it is found that the current model performs poorly on high-difficulty problems. Researchers can then specifically enhance the model’s reasoning abilities or improve its performance on complex data. Additionally, difficulty classification enables researchers to design test and training sets more rationally, making them more challenging or balanced. This not only tests the models' limits but also enhances their robustness.

\subsection{Performance of Models on Tasks of Varying Difficulty}

Our study tested simple route planning tasks to assess the model's performance in basic pathfinding and reasoning capabilities. However, despite the simplicity of the task, the model's performance on more challenging tasks in this category was still suboptimal (see Table~\ref{tab:t3}). We found that the model's performance significantly declined when handling complex route planning problems, particularly in scenarios involving larger grids or more obstacles. Similar observations were made in the work of Mohamed Aghzal. His research indicated that, in complex route planning tasks, particularly when the number of obstacles increased, gpt-4 (with CoT prompting) struggled significantly, suggesting that it still has limitations in complex spatial reasoning\cite{aghzal2024large}. His study further noted that when tasks involved long-distance planning, models like gpt-4 exhibited decreased performance, particularly when the target was distant, indicating deficiencies in long-duration, cross-temporal reasoning. Furthermore, our research suggests that in spatial reasoning tasks, the difficulty is primarily determined by the complexity of the spatial description and the number of elements involved. More elements imply greater complexity and, consequently, greater difficulty. A similar conclusion was reached in the work of Yamada, who noted that in spatial reasoning tasks, difficulty is not only related to the type of spatial structure but also closely tied to the number of navigation steps\cite{yamada2023evaluating}. In particular, when dealing with complex grid structures, the model's performance was significantly influenced by both the spatial structure and the number of navigation steps. For example, gpt-4 performed best on square grid structures and worst on hexagonal grids, highlighting that different spatial layouts present distinct challenges to the model.

To further investigate the differences between models on medium and difficult tasks, we randomly selected moderate and challenging problems from various task categories and compared the performance of moonshot-v1-8k and gpt-4o (see Table~\ref{tab:t6}). In these more complex tasks, we observed that gpt-4o consistently outperformed moonshot-v1-8k in handling these problems. This suggests that while moonshot-v1-8k performs reasonably well on simpler tasks, its performance significantly lags when confronted with complex spatial reasoning or route planning problems.

We believe that this difference may be attributed to several factors: first, gpt-4o's training parameters may be more finely tuned, particularly in terms of reasoning and optimization for complex tasks; second, gpt-4o's training framework may be more advanced, enabling it to better handle complex route planning tasks; third, gpt-4o may have been trained on a larger and higher-quality dataset, which enhances its ability to solve complex problems. Consequently, whether in route planning or spatial reasoning, the model's performance on complex reasoning tasks is constrained, which aligns with the findings of Aghzal and Yamada. Future research should address these challenges to enhance the model's reasoning capabilities in more complex scenarios.

\begin{table}[htbp]
\centering
\caption{\label{tab:t6}Performance of moonshot-v1-8k and gpt-4o on tasks of different difficulty,(\(\checkmark\)) represents a correct answer, and (\(\times\)) represents an incorrect answer.}
\renewcommand{\arraystretch}{1.2} 
\resizebox{\textwidth}{!}{%
\begin{tabular}{p{8cm} p{4cm} p{4cm} p{4cm}}
\toprule
\textbf{Questions (difficulty level)} & \textbf{moonshot-v1-8k} & \textbf{gpt-4o} & \textbf{Truth} \\ 
\midrule
You are in a 4 by 4 two-dimensional array. You have to avoid some obstacles: (1,0). From (0,0) to (3,3). Just use ``up, down, left, right" to indicate your route. Your output is formatted as: path: (medium) & up, up, up, right, down, down, right, right (\(\times\)) & path: right, right, right, down, down, down (\(\checkmark\)) & right, right, right, down, down, down \\ 
\cmidrule(lr){1-4}
You are in a 7 by 7 two-dimensional array. You must avoid some obstacles: (6,2). From (3,3) to (0,5). Just use ``up, down, left, right" to indicate your route. Your output is formatted as: path: (difficult) & up, up, up, right, right, right (\(\times\)) & path: up, up, up, right, right (\(\checkmark\)) & up, up, up, right, right \\ 
\cmidrule(lr){1-4}
You are in a 7 by 7 two-dimensional array. You must avoid some obstacles: (4,3). From (0,0) to (6,2). Just use ``up, down, left, right" to indicate your route. Your output is formatted as: path: (difficult) & up, up, right, right, right, down, down, right (\(\times\)) & path: right, right, right, down, down, down, down, down, down, down, left, left (\(\times\)) & right, right, down, down, down, down, down, down, down \\ 
\cmidrule(lr){1-4}
You have been given a pointy-topped regular hexagonal tile map consisting of 1 tile. Starting from a vertex, you will move along the edges of the tile. Initially, you are positioned at the bottom right corner of the map, where you find a husky. You move up by one step, where you find a giant panda. You move up-left by one step, where you find a moped. You move down-left by one step, where you find a crane (machine). You move down by one step, where you find a scooter. You move down-right by one step, where you find a hoop skirt. You move up-right by one step. What will you find? (medium) & giant panda (\(\times\)) & You will find a husky. (\(\checkmark\)) & husky \\ 
\cmidrule(lr){1-4}
You have been given a circular grid consisting of 8 connected dots. Starting from a vertex, you will move along the edges of the circular grid. Initially, you are positioned on the dot that's located at the top of the grid, where you find a wing. You move around the ring by 5 steps in a counter-clockwise direction, where you find a starfish. You move around the ring by 8 steps in a clockwise direction. What will you find? (difficult) & Wing (\(\times\)) & You will find a wing. (\(\times\)) & starfish \\ 
\cmidrule(lr){1-4}
You have been given a 3 by 3 square grid. Starting from a vertex, you will move along the edges of the grid. Initially, you are positioned at the bottom right corner of the grid, where you find a power drill. You move left by one step, where you find a sock. You move up by one step, where you find a Dalmatian. You move up by one step, where you find a jeep. You move right by one step, where you find an hourglass. You move down by one step, where you find a slip-on shoe. You move left by one step. What will you find? (difficult) & You will find a power drill. (\(\times\)) & You will find a jeep. (\(\times\)) & Dalmatian \\ 
\bottomrule
\end{tabular}%
}
\end{table}

\subsection{The Capabilities of Multimodal Models in Geographic Tasks}

In the ``Example Analysis" section, we identified significant differences in the performance of LLMs when executing NL2API Mapping tasks. To illustrate these differences, we compared gpt-4o and gpt-4-turbo-2024-04-09 using a specific task: generating a Mapbox map link for downtown San Francisco. While seemingly straightforward, this task tested various aspects of the models’ geographic information processing capabilities. To better understand these variations, we analyzed the key differences in the models’ responses from three perspectives:

Firstly, in terms of semantic understanding of geographic concepts, gpt-4o demonstrated a more precise interpretation of ``downtown San Francisco". The generated map focused on the central business district, particularly around Union Square and the Financial District. This aligns closely with the traditional geographic definition of downtown in urban planning and local contexts. Secondly, regarding technical implementation through the Mapbox API, both models utilized the same Mapbox GL JS framework and style definitions. Differences in map detail levels were primarily controlled by the zoom level parameter in the API calls, rather than direct manipulation of map content. This is consistent with the functionality of Mapbox’s vector tile system, where the visibility of different features (e.g., labels and points of interest) is predetermined by style definitions at various zoom levels. Finally, in translating natural language requirements into technical parameters, gpt-4o exhibited greater precision in parameter selection. It chose a more appropriate zoom level and central coordinates, accurately representing the concept of ``downtown". In contrast, while gpt-4-turbo-2024-04-09 generated a technically valid map, its broader view included residential areas west of the downtown core, indicating a less focused understanding of the task requirements.

These differences did not stem from the models’ direct understanding of unstructured map data (as the maps were generated through structured API calls) but rather from their varying capabilities in: 

\begin{itemize}[leftmargin=2em, rightmargin=2em]
\item Accurately interpreting geographic concepts and spatial relationships
\item Applying appropriate technical parameters within the constraints of the mapping API 
\item Bridging the gap between natural language requirements and technical implementation
\end{itemize}

This case study highlights that even in seemingly simple geographic tasks, the models’ approach to converting semantic input into technical implementation can significantly impact the outcome. While both models generated valid map links, their differing methods of parameter selection and geographic concept interpretation resulted in markedly different spatial representations of the same area. This analysis provides an important insight into evaluating language models for geographic tasks: the ability to accurately translate geographic concepts into technical parameters is as crucial as interpreting unstructured map data. Although the present case study primarily demonstrates the former through API interactions, both skills are foundational for comprehensive geographic understanding. Future GIS applications of LLMs must excel in both precise parameter selection and spatial-semantic interpretation of unstructured geographic information, underscoring the dual challenge for their development.

\subsection{Enhancing Performance of LLMs on Complex Spatial Tasks with Prompt Strategies}

LLMs are widely recognized for significantly improving output quality by optimizing input prompts. However, our research reveals significant differences in how different models respond to prompts. In this experiment, we defined NL2API Mapping, simple route planning, and spatial understanding as complex spatial tasks. The majority of the test results for these tasks indicated that models like gpt-4o exhibited greater ``versatility", improving performance across various prompt strategies. This was particularly evident in reasoning tasks, where the use of chain-of-thought (CoT) and zero-shot CoT significantly enhanced performance\cite{wei2022chain}. In contrast, models like glm-4 showed a different pattern of responses in spatial understanding tasks, with each prompt strategy seemingly leading to a decline in performance. This difference may be due to several factors: 1) Differences in architectural design between glm-4 and models like gpt-4o may lead to varying performance when handling prompt strategies\cite{glm2024chatglm}\cite{openai2024gpt4o}; 2) glm-4 may be relatively weaker in basic spatial understanding and reasoning abilities compared to models like gpt-4o, meaning that even with optimized prompt strategies, its performance may not improve significantly and could even degrade due to information overload or improper processing; 3) glm-4 may face bottlenecks when handling complex tasks, and the introduction of prompt strategies could further increase task complexity, impacting performance. Therefore, to fully realize the potential of different LLMs, it is essential to carefully design prompts for specific tasks.

\subsection{Limitations of the Study}

To accurately assess and compare the performance of different LLMs on spatial tasks, our study had two core requirements: 1) establishing an independent and diverse spatial task dataset, and 2) designing a comprehensive testing methodology, including the refinement of variable design and evaluation strategies. However, the spatial task categories used in this study are not comprehensive. Previous research has conducted more detailed analyses of other spatial task categories, such as point of interest (POI) recommendation, vector data analysis\cite{zhang2024geogpt}\cite{chen2024mapgpt}, and map analysis. Additionally, our study primarily focused on text-based spatial tasks, while real-world applications of spatial tasks often involve multimodal elements such as charts and maps. Since some models do not support images as input, our tests did not include image-related tasks. In the second round of prompt optimization testing, due to certain limitations in the spatial task outputs, we were unable to adopt more complex prompt strategies, such as Self-consistency: sampling multiple solutions and then performing majority voting\cite{wang2023self}, using complex chains instead of simple chains as context examples\cite{fu2022complexity}, and decomposing complex tasks into simpler tasks to solve them sequentially\cite{khot2023decomposed}. Therefore, introducing multimodal test data and designing more comprehensive testing methodologies are key directions for future work. 

\section{Conclusions}

In our study, we designed a comprehensive spatial task dataset, including various task categories such as GIS concepts and simple route planning, and systematically evaluated the performance of different LLMs (e.g., gpt-4o, glm-4) on these tasks. We then classified the dataset by difficulty levels to further evaluate the models' performance in handling problems of varying difficulty. We also tested various prompt strategies (including One-shot, Chain of Thought, etc.) to assess their effectiveness in enhancing model performance on complex spatial tasks. The results indicate that different large models have their own strengths and limitations in spatial tasks, and their performance generally decreases with increasing task complexity. Specifically, most models did not perform well on tasks requiring high-level reasoning, such as simple route planning. Additionally, there are differences in the sensitivity of different models to prompt strategies. The effectiveness of prompt strategies on Moontshot and GLM models is generally lower than on OpenAI and Gemini models, indicating the need for more tailored prompt strategies to optimize performance for specific model architectures. Overall, our study established a systematic and comprehensive benchmark to test the performance of different LLMs on various spatial tasks. This allows us to assess which models are better suited for specific spatial tasks and provides a scientific basis for targeted performance improvements of future models. 

\section{Findings}

This work is supported by the Natural Science Foundation of Zhejiang Province [grant
number LGG22D010001, LQ19D010011].

\section{Notes on contributors}

\textbf{Liuchang Xu}

\textbf{Liuchang Xu} received the B.S. degree in applied psychology from Zhejiang University, China, in 2015, and the Ph.D. degree in Remote sensing and Geographic
Information System from Zhejiang University, China, in 2020. He is currently an
associate Professor at the School of Mathematics and Computer Science, Zhejiang
A\&F University, China. His research interests include Spatio-temporal Analytics,
artificial intelligence, Urban Computing and Big Data Mining.

\textbf{Shuo Zhao}

\textbf{Shuo Zhao} is currently pursuing his M.S. at the Zhejiang A\&F University. His research interests include Geographic Information Processing and Artificial Intelligence.

\textbf{Qingming Lin}

\textbf{Qingming Lin} is currently pursuing his M.S. at the Zhejiang A\&F University. His research interests include Geographic Information Processing and Artificial Intelligence.

\textbf{Luyao Chen}

\textbf{Luyao Chen} is currently pursuing her M.S. at Zhejiang A\&F University. Her research interests include Remote Sensing and Artificial Intelligence.

\textbf{Qianqian Luo}

\textbf{Qianqian Luo} is currently pursuing her M.S. at Zhejiang A\&F University. Her research interests include Remote Sensing and Artificial Intelligence.

\textbf{Sensen Wu}

\textbf{Sensen Wu} received the Ph.D. degree in cartography and geographic information
systems from Zhejiang University, Hangzhou, China, in 2018. He is currently working
as an Associate Professor with the School of Earth Sciences, Zhejiang University. His
research interests include spatial-temporal analysis, remote sensing, and deep learning.

\textbf{Xinyue Ye}

\textbf{Xinyue Ye} is the Harold Adams Endowed Professor in Urban Informatics and Stellar Faculty Provost Target Hire at Texas A\&M University (TAMU). His current research is centered on urban digital twins and precision public health, emphasizing real-time 3D modeling and AI-enabled participatory planning, as well as urban climate science, with a focus on downscaling climate data to the built environment scale and its relevance to human mobility.

\textbf{Hailin Feng}

\textbf{Hailin Feng}, professor, doctoral supervisor. Graduated from the University of Science and Technology of China with a doctorate in computer application technology. Currently serving as the vice dean of the School of Mathematics and Computer Science at Zhejiang A\&F University (in charge of work). He is a young and middle-aged academic leader in Zhejiang Province’s colleges and universities, a key academic leader in Zhejiang Province, a member of the Fault-Tolerant Computing Committee of the China Computer Society, and a member of the Smart Agriculture Committee of the Chinese Automation Society.

\textbf{Zhenhong Du}

\textbf{Zhenhong Du}, professor, doctoral supervisor, winner of the National Science Fund
for Distinguished Young Scholars, Qiushi Distinguished Professor of Zhejiang University. Dean of the School of Earth Sciences, Zhejiang University. His research interests
include Spatio-temporal big data Mining, artificial intelligence.

\section{ORCID}

Liuchang Xu \href{https://orcid.org/0000-0001-7635-7266}{https://orcid.org/0000-0001-7635-7266}

\section{Data availability statement}

The data that support the findings of this study are available with the identifier(s) at the private link: \href{https://figshare.com/s/be55522f22bf761cfcab}{https://figshare.com/s/be55522f22bf761cfcab}

\section{Declarations}

\textbf{Conflict of interest} No potential conflict of interests was reported by the authors. 

\textbf{Consent to participate} Not applicable. 

\textbf{Consent for publication} Not applicable. 

\textbf{Ethics approval} Not applicable.

\begin{refcontext}[sorting = none]
\printbibliography
\end{refcontext}
\appendix
\section{Proofs}

\begin{figure}[H]
\centering
\includegraphics[width=0.9\linewidth]{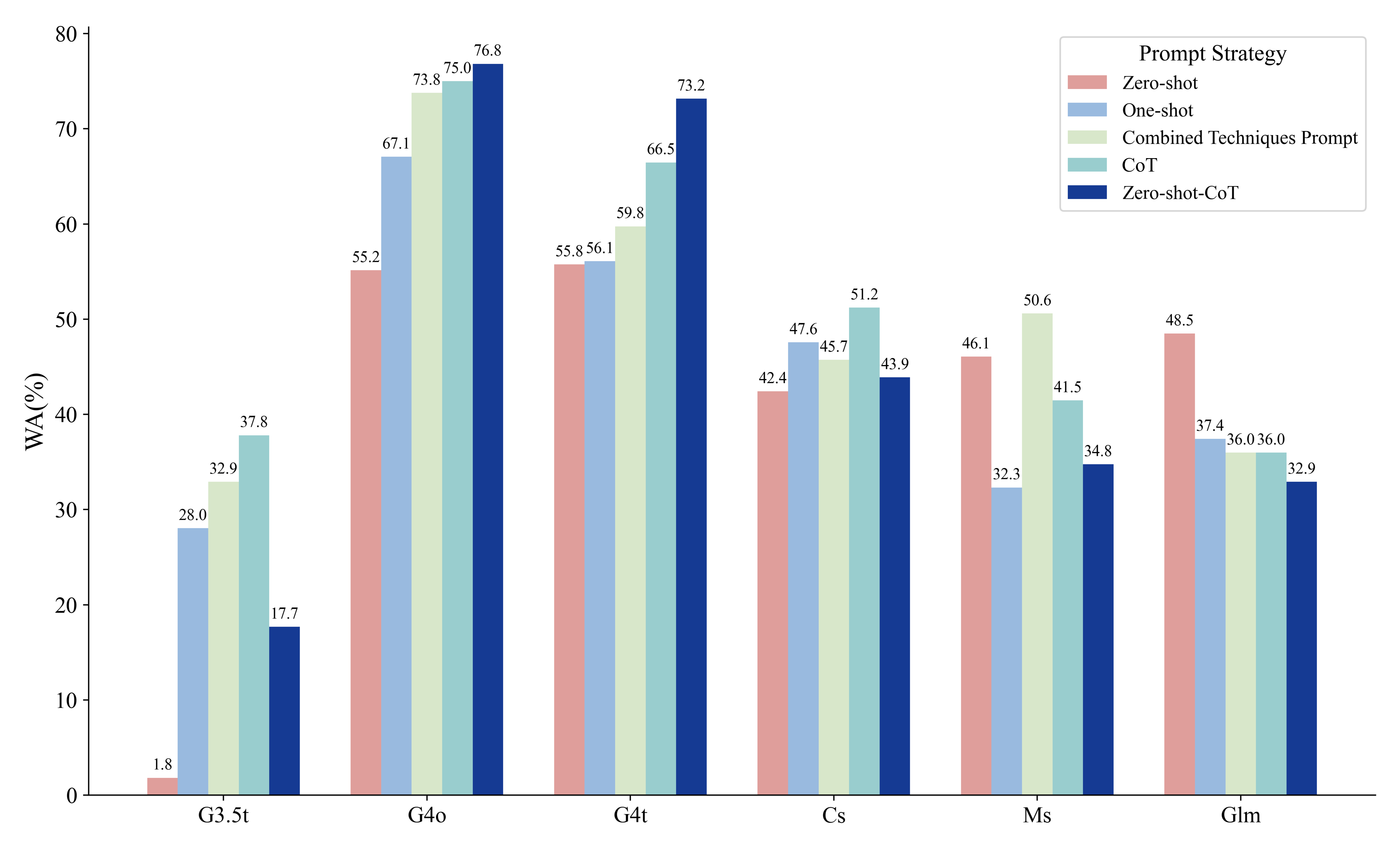}
\caption{\label{figure-A1:PNG} Prompt optimization test results for spatial understanding task}
\end{figure}

\begin{figure}[H]
\centering
\includegraphics[width=0.9\linewidth]{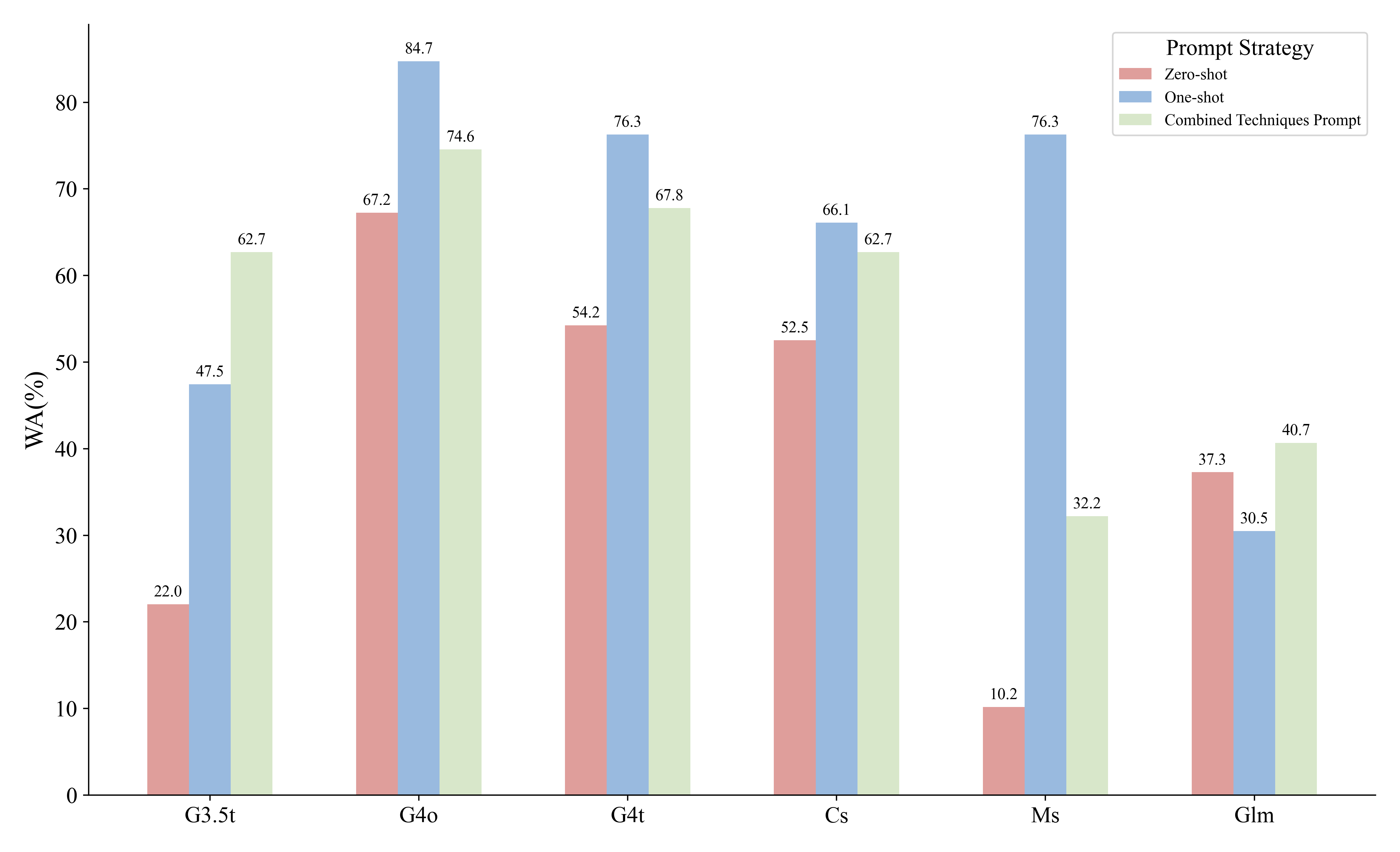}
\caption{\label{figure-A2:PNG} Prompt optimization test results for NL2API Mapping task}
\end{figure}

\end{document}